\documentclass[11pt]{article}

\usepackage[a4paper, margin=1in]{geometry}
\usepackage[utf8]{inputenc}
\usepackage[T1]{fontenc}
\usepackage{lmodern}
\usepackage{microtype}

\usepackage{amsmath, amssymb, amsfonts, amsthm}
\usepackage{xfrac}
\usepackage{dsfont}
\usepackage[mathscr]{eucal}

\usepackage{xcolor}
\usepackage{graphicx}
\usepackage{booktabs}
\usepackage{multicol}
\usepackage{algorithm}
\usepackage[numbers,sort&compress]{natbib}
\usepackage{hyperref}
\hypersetup{
  colorlinks=true,
  linkcolor={red!70!black},
  citecolor={blue!70!black},
  urlcolor={red!70!black}
}

\theoremstyle{plain}
\newtheorem{result}{Result}

\DeclareMathOperator{\var}{Var}
\DeclareMathOperator{\tr}{Tr}
\DeclareMathOperator*{\argmax}{arg\,max}
\DeclareMathOperator*{\argmin}{arg\,min}

\title{High-Dimensional Theory of LoRA Fine-Tuning \\ in a Solvable Attention Model}

\author{
\includegraphics[height=0.79em]{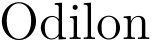} Duranthon$^1$,
Fabrizio Boncoraglio$^1$,
Lenka Zdeborov\'a$^1$
}

\date{
\small
$^1$ Statistical Physics of Computation Laboratory,
\\
École Polytechnique Fédérale de Lausanne (EPFL)
CH-1015 Lausanne
}

\begin{document}
\maketitle

\begin{abstract}
  We develop a high-dimensional statistical theory of low-rank adaptation (LoRA) in attention models, capturing the interplay between pre-training and fine-tuning. We introduce a solvable framework in which a single-head attention layer is first pre-trained on a data-abundant task and subsequently adapted via a rank-one LoRA update on limited data. In the high-dimensional limit, both stages admit a sharp asymptotic characterization in terms of a finite set of order parameters, yielding explicit predictions for test errors and representation alignment. Our analysis shows that the impact of pre-training on LoRA is summarized by an effective noise term, from which we derive prescriptions for the optimal pre-training procedure. We also demonstrate a regime with a mismatch between the value of the test error and representation quality, and propose an application of our theory to active fine-tuning.
\end{abstract}

\section{Introduction}

Large pre-trained attention models are rarely retrained from scratch for each new task.
Instead, they are adapted to downstream data through parameter-efficient fine-tuning methods, which keep most pre-trained weights frozen and optimize only a small number of task-specific parameters. Among these methods, low-rank adaptation (LoRA) has become a standard approach: it injects trainable low-rank updates into the weight matrices of a frozen transformer, often matching or approaching full fine-tuning performance while substantially reducing memory, storage, and optimization costs~\cite{hu2022lora}. This empirical success is part of a broader line of parameter-efficient adaptation methods, including adapters~\cite{houlsby2019parameter}, prefix tuning~\cite{li2021prefix} and QLoRA~\cite{dettmers2023qlora}. A recurring empirical observation is that fine-tuned adaptation appears to live in a much lower-dimensional space than pre-training, as also suggested by intrinsic-dimension studies of language-model fine-tuning~\cite{aghajanyan2021intrinsic}. 

Recent theoretical work has begun to clarify some aspects of LoRA. Existing results mainly address its expressivity, optimization geometry or dynamics \cite{zeng24expressivityLora, zhang25oneStepLora,kim25convergenceLora,xu25dynamicsLora,liang25robustnessLora,mu25convergenceLora,nwemadji26lora,steele26robustnessLora}. 
Closer to our work, \cite{Kratsios25loraGenBound} derives generalization bounds for asymmetric LoRA, where one factor of the LoRA is frozen at random and only the other is trained, while \cite{jang25loraNTK} provides an NTK-based analysis that characterizes optimization and generalization in a lazy regime. These works shed light on important facets of LoRA, but leave open the question of a quantitative statistical theory for adaptation after pre-training. In particular, a characterization of how fine-tuning performance depends on both (large) pre-training and (small) fine-tuning sample sizes and pre-training quality is still lacking.

In this work we develop such a theory in a solvable high-dimensional model of LoRA fine-tuning for attention. We consider a single-head tied-attention layer that is first pre-trained on an extensive-rank attention-indexed (AIM) target~\cite{boncoraglio2025bayes,boncoraglio2026singleheadattentionhighdimensions} with a large number of samples, and is then adapted to a downstream/fine-tuning task by only learning a rank-one LoRA component from a much smaller dataset. This yields a minimal setting in which pre-training a large model, freezing, and low-rank adaptation coexist in a form that remains analytically tractable. Our main result is a sharp asymptotic characterization of the LoRA-fine-tuned estimator. In the high-dimensional limit, both the pre-training stage and the fine-tuning stage are described by a finite set of scalar order parameters, from which we obtain explicit predictions for the pre-training test error, the fine-tuning test error, and the overlaps with the ground truth. This provides, to our knowledge, the first tight high-dimensional learning-curve theory for LoRA fine-tuning after attention pre-training.

\paragraph{Contributions.}
Our main contributions are:
\begin{enumerate}
    \item We introduce a solvable high-dimensional setting of LoRA fine-tuning in attention, with a full-rank pre-training stage on a large dataset and a rank-one fine-tuning on a much more limited dataset.
    \item We derive a sharp asymptotic characterization of the LoRA test error and reconstruction overlap in terms of finite-dimensional order parameters. Technically, we rely on the versatility of the high-dimensional analysis to merge independent results for the pre-training and the fine-tuning task into a joint characterization.   
    \item We identify an effective-noise mechanism which quantifies how the quality of pre-training objective controls fine-tuning performance. This single parameter allows us to derive quantitative prescriptions for the optimal pre-training procedure and an idealized active-learning rule for fine-tuning.
    \item We identify regimes in which there is a mismatch at the fine-tuning stage between good test error and good reconstruction overlap thanks to memorization of the pre-training data.
\end{enumerate}

\paragraph{Further related work.}
Our work fits into a growing line of exact high-dimensional theories for feature learning, from multi-index models~\cite{benarous2021online,damian2023smoothing,arnaboldi2023dimensionless,berthier2024learning} to sequence multi-index models \cite{cui2025high,cui2024highdimensionallearning,troiani2025fundamental,arnaboldi2025asymptotics,duranthon25slr} and attention architectures~\cite{boncoraglio2025bayes,boncoraglio2026singleheadattentionhighdimensions}. The specific contribution of the present paper is to bridge extensive-rank pre-training and low-rank adaptation, and thereby provide a statistical theory of LoRA in the feature-learning regime rather than in a kernel or purely optimization-based limit.

\paragraph{Notations.}
For a positive integer $n$ we write $[n]=\{1,\ldots,n\}$; $I_n$ the $n\times n$ identity matrix; $\mathscr S(n)$ the set of symmetric $n\times n$ matrices. $||\cdot||_F$ is the Frobenius norm of a matrix.
We denote a Gaussian law centered at $\omega$ with covariance $V$ as $\mathcal N(\omega,V)$.

\section{A two-stage model for pre-training and LoRA fine-tuning}
\label{sec:2}

We consider sequences of $T$ tokens of dimension $D$ for a sequence-to-sequence task. We have access to a large dataset of pre-training samples and to a smaller dataset for fine-tuning. More precisely, we consider a set of $N$ pre-training input and output sequences $\mathcal S=(X_\mu\in\mathbb R^{T\times D}, y_\mu\in\mathbb R^{T\times D})_{\mu\in[N]}$ and a set of $N'$ fine-tuning input and output sequences $\mathcal S'=(X'_\mu\in\mathbb R^{T\times D}, y'_\mu\in\mathbb R^{T\times D})_{\mu\in[N']}$.

\textbf{Model Architecture:} we consider an attention layer with tied keys and queries and identity values, parameterized by the weights $W\in\mathbb R^{D\times P}$, with $P$ the rank of $W$, and $w\in\mathbb R^D$. $W$ is the large-rank part of the weights that are pre-trained, while $w$ is the low-rank part that is fine-tuned. Given a sample $X,y$ from $\mathcal S$ or $\mathcal S'$, the attention predicts the sequence
\begin{align}
 \hat y(X;W,w) = \sigma(A-\mathbb E_XA)X\, , \quad \quad
 A = D^{-1}X\left(P^{-\sfrac{1}{2}}WW^\top + ww^\top\right)X^\top\, ,
\end{align}
where $\sigma:\mathbb R^{T\times T}\to\mathbb R^{T\times T}$ is the activation function. In the following we will consider it to be the row-wise softmax or the rescaled identity $\sigma:x\mapsto x/T$. $A$ is the pre-activated attention matrix. $\mathbb E_X$ is the empirical expectation over the sequences of the set $\mathcal S$ or $\mathcal S'$ and the term $-\mathbb E_XA$ is a batch normalization. The scaling with respect to $D$ and $P$ is taken so that all the elements of $A$ are of order one, and $W$ and $w$ contribute comparably to $A$.

\textbf{Pre-training:} we first pre-train the attention by minimizing the following empirical loss
\begin{align}
\mathcal L(W)=\sum_\mu^ND^{-1}||\hat y(X_\mu;W,0)-y_\mu||_F^2\ .
\end{align}
We fix $\hat W\in\argmin_W(\mathcal L+\mathscr R)$ a minimizer of the loss, with $\mathscr R(W)=\lambda||W||_F^2$ a regularization. 

\textbf{LoRA fine-tuning:} we then fine-tune the attention on the loss
\begin{align}
\mathcal L'(w)=\sum_\mu^{N'}D^{-1}||\hat y(X'_\mu;\hat W,w)-y'_\mu||_F^2
\end{align}
taking $\hat w\in\argmin_w(\mathcal L'+\mathscr R')$, with $\mathscr R'(w)=\lambda'||w||_2^2$. $\lambda,\lambda'>0$ are the regularization strengths.

\textbf{Measured performances:} we are interested in the test errors of the attention with weights $\hat W, \hat w$ on test pre-training and fine-tuning samples
\begin{align} 
\label{eq:testError}
\mathcal{E}(\hat W) &= \mathbb{E}_{(X,y)\sim\mathcal P}\left[D^{-1}||\hat y(X;\hat W,0)-y||_F^2 \,\big|\, \mathcal{S}\right] \\
\mathcal{E}'(\hat W, \hat w) &= \mathbb{E}_{(X',y')\sim\mathcal P'}\left[D^{-1}||\hat y(X';\hat W,\hat w)-y'||_F^2 \,\big|\, \mathcal{S}, \mathcal{S}'\right] 
\end{align}
We will also consider the reconstruction error of the LoRA task measured as the cosine similarity (or \emph{overlap}) between the learned weights and the ground truth $o_w = \sfrac{|\hat w^\top w^*|}{\|\hat w\|_2\|w^*\|_2}$.

For a generic pre-training and fine-tuning dataset, the two-stage optimization above is too complex to admit a sharp analytical description. To make the problem tractable while preserving the key ingredients of pre-training followed by low-rank adaptation, we consider the data to be synthetic, generated by a target of the same type of the analyzed architecture. 

\textbf{Probabilistic model for synthetic data:} 
let $P_0$ be the rank of the target and take $W^*\in\mathbb R^{D\times P_0}$, $w^*\in\mathbb R^D$ common to all samples, with $W^*_{i,j}\sim\mathcal N(0,1)$ and $w^*_i\sim\mathcal N(0,1)$ for all $i\in[D], j\in[P_0]$. We consider two possible scenarios, that we distinguish by the index $\mathrm e\in\{0,1\}$:\\
\textcolor{white}{--}\emph{i) fine-tuning on different data:} in this case the sequences $X$ and $X'$ of the pre-training and fine-tuning train sets are different and we draw them independently. We set $\mathrm e=0$.\\
\textcolor{white}{--}\emph{ii) fine-tuning on reused sequences:} in this case the input sequences are shared between the pre-training and fine-tuning, while the output sequences correspond to different tasks. We set $\mathrm e=1$.\\
We set $\mathcal P$ and $\mathcal P'$ to be the pre-training and fine-tuning distributions of the samples. They are jointly defined by
\begin{align}
(\mathcal P, \mathcal P') : \left\{\begin{matrix}
X\sim \mathcal P_0 \\
\xi\sim\mathcal{P}_\Delta \\
A^* = D^{-1}P_0^{-\sfrac{1}{2}}XW^*W^{*\top}X^\top \\
y = \sigma(A^*-\mathbb E_XA^*+\xi)X
\end{matrix}\right.\ ,
\left\{\begin{matrix}
X'\sim\mathrm e\delta_{X}+(1-\mathrm e)\mathcal P_0\\
\xi'\sim\mathrm e\delta_{\xi}+(1-\mathrm e)\mathcal{P}_\Delta \\
{A'}^* = D^{-1}X'\left(P_0^{-\sfrac{1}{2}}W^*W^{*\top}+w^*w^{*\top}\right){X'}^\top\\
y' = \sigma({A'}^*-\mathbb E_{X'}{A'}^*+\xi')X'
\end{matrix}\right.
\end{align}
where $(\mathcal P_0)_{a,i}=\mathcal N(0,1)$ for all $a\in[T], i\in[D]$; $\xi,\xi'\in\mathscr S(T)$ are symmetric noise with strength $\Delta\geq 0$ and law $(\mathcal{P}_\Delta)_{a,b}=\mathcal N(0,\Delta/(2-\delta_{a,b}))$ for all $a\leq b$. The factor $1/(2-\delta_{a,b})$ is to take into account the symmetrization. $\mathbb E_X$ and $\mathbb E_{X'}$ are expectations over the sequences. For $\mathrm e=0$ the two distributions factorize over the samples, while for $\mathrm e=1$ we match all the randomness. $\mathcal S$ and $\mathcal S'$ are then constructed by independently drawing $N'$ samples from $(\mathcal P, \mathcal P')$ and the remaining $N-N'$ from $\mathcal P$.

\textbf{Asymptotic limit:} we consider a high-dimensional limit $D\to\infty$ together with
\begin{align}
& \frac{P_0}{D}\to\kappa_0=\Theta(1)\ ,\quad \frac{P}{D}\to\kappa=\Theta(1)\ ,\quad \frac{N}{D^2}\to\alpha=\Theta(1)\ ,\quad \frac{N'}{D}\to\alpha'=\Theta(1)
\end{align}
with $\kappa \ge 1$ and a finite number of tokens $T=\Theta(1)$. This means that, for the pre-training, we consider $W$ and $W^*$ to have extensive rank and we have access to a quadratically large number of samples; while for the fine-tuning we consider $w$ and $w^*$ low-rank (rank one) and we only have access to a limited linear number of samples. We emphasize that $N\gg N'$, which corresponds to the common case where the fine-tuning samples are limited compared to the pre-training samples.

\textbf{Comments and limitations:}
the pre-training and fine-tuning tasks share a common extensive-rank component, which must first be learned during pre-training and then transferred to the downstream/fine-tuning task. The additive noise $\xi$ models residual interactions and task variability that are not captured by the attention architecture, and therefore allows for a controlled mismatch between the target and the student. We focus on Gaussian input sequences for analytical tractability, but, as in related high-dimensional models \cite{montanari2022universality,dandi2023universality,dudeja2023universality,lu2025equivalence}, we expect the resulting characterization to be robust beyond the strictly Gaussian setting. Finally, we restrict here to tied keys and queries and to a rank-one LoRA update; these choices yield the simplest setting in which the interplay between pre-training and low-rank adaptation can be analyzed exactly. Extensions to untied attention and higher-rank updates can be directly obtained following \cite{boncoraglio2026singleheadattentionhighdimensions} and \cite{cui2025high}. The assumption $\kappa\geq1$ avoids a finite-width bottleneck in the pre-training stage. The case $\kappa<1$ can be treated by including the corresponding rank constraint, but we do not consider it here in order to focus on the transfer from the frozen extensive-rank component to the rank-one LoRA update.

\section{Asymptotic characterization of the LoRA}
\label{sec:characterization}
This section describes the main technical contribution of the paper that consists of using the existing analysis of the pre-training-only model \cite{boncoraglio2026singleheadattentionhighdimensions} with the rank-one fine-tuning. In absence of pre-training the fine-tuning task corresponds to the sequence multi-index model \cite{cui2025high}. Due to the versatility of the high-dimensional analysis presented in these two lines of work, one is able to adapt these two results into a joint characterization of the pre-training and fine-tuning as derived in Appendix \ref{App:A}. We describe the resulting characterization with details given in \ref{App:A}.

\textbf{Order parameters:} in the high-dimensional limit, the observables of the trained attention $\mathcal L$, $\mathcal L'$, $\mathcal{E}$, $\mathcal{E}'$ and $o_w$ can be solely described in terms of the following scalar order parameters (or \emph{sufficient statistics}), setting $S^*=P_0^{-\sfrac{1}{2}}W^*W^{*\top}$ and $\hat S=P^{-\sfrac{1}{2}}\hat W\hat W^{\top}$:
\begin{align}
\label{eq:order_parameters}
Q_0 &= D^{-2}\tr S^*S^* & Q &= D^{-2}\tr \hat S\hat S & M &= D^{-2}\tr \hat SS^* \\
q_0 &= D^{-1}w^{*\top}w^* & q &= D^{-1}\hat w^{\top}\hat w & m &= D^{-1}\hat w^{\top}w^*
\end{align}
$Q$ and $q$ quantify the norms of the weights while $M$ and $m$ quantify how much the weights are aligned with the ground truth. In the high-dimensional limit we have $Q_0=1+\kappa_0$ and $q_0=1$. We moreover introduce the order parameters $V,v\in\mathbb R_+$, that quantify the inverse curvature of the train loss at the minimizer. Large $V$ or large $v$ means that the loss is flat, thus the minimizer is sensitive to perturbations and likely overfits data.

\subsection{Effective low-dimensional pre-activations}

A standard consequence of high-dimensional concentration in teacher--student models is that the behavior of the predictor can be reduced to a finite-dimensional effective problem. In the present setting, all observables of interest can be expressed in terms of the joint distribution of a small number of scalar pre-activations associated with the target and the trained attention. We therefore introduce these effective variables first.
For a sequence $X$ let the following be the pre-activations of the target and the trained attention.
\begin{align}
\label{eq:preactivations}
z^* &= A^*-\mathbb E_XA^*+\xi \qquad\mathrm{where}\; A^* = D^{-1}P_0^{-\sfrac{1}{2}}XW^*W^{*\top}X^\top \\
z &= A-\mathbb E_XA \qquad\mathrm{where}\; A = D^{-1}P^{-\sfrac{1}{2}}X\hat W\hat W^{\top}X^\top \\
\chi^* &= D^{-\sfrac{1}{2}}Xw^* \qquad\mathrm{and}\qquad \chi = D^{-\sfrac{1}{2}}X\hat w
\end{align}

We characterize the distribution of these pre-activations firstly for $X$ being a test sequence.
\begin{result}[Test samples]
\label{res:testPreAct}
Consider $X,\xi$ being distributed from $\mathcal P$, independent of the samples of $\mathcal S$. Then the pre-activations $(z^*,z)\in(\mathscr S(T))^2$ follow the distribution $\mathcal Q$ defined for all $a\leq b\in[T]$ by
\begin{align}
\left(\begin{smallmatrix}z^*_{a,b} \\ z_{a,b}\end{smallmatrix}\right) =\left(\begin{smallmatrix}z^*_{b,a} \\ z_{b,a}\end{smallmatrix}\right) \sim \mathcal Q_{ab}=\mathcal N\left(0,(1+\delta_{a,b})\left(\begin{smallmatrix}\Delta/2+Q_0 & M \\ M & Q\end{smallmatrix}\right)\right)\ .
\end{align}
Consider $X$ being distributed from $\mathcal P_0$, independent of the samples of $\mathcal S$ and $\mathcal S'$. Then the pre-activations $(\chi^*,\chi)\in(\mathbb R^T)^2$ follow the distribution $\mathcal Q'$ defined for all $a\in[T]$ by
\begin{align}
\left(\begin{smallmatrix}\chi^*_{a} \\ \chi_{a}\end{smallmatrix}\right) \sim \mathcal Q'_{a}=\mathcal N\left(0,\left(\begin{smallmatrix}q_0 & m \\ m & q\end{smallmatrix}\right)\right)\ .
\end{align}
\end{result}

We then turn to train samples. Because of the correlations between the trained weights and the train sequences, the distribution of the pre-activations is more involved. We introduce the proximal operators, for the pre-training and the fine-tuning. For symmetric $\tilde z\in\mathscr S(T)$ and for $\tilde\chi\in\mathbb R^T$ define the following scalar potentials
\begin{align}
\Psi(z; z^*,\tilde z) &= -||\sigma(z)-\sigma(z^*)||_F^2-\sum_{a\leq b}^T\frac{1}{2V(1+\delta_{a,b})}(\tilde z_{a,b}-z_{a,b})^2 \\
\psi(\chi; \chi^*,\tilde\chi,z^*,z) &= -||\sigma(z+\chi\chi^\top-qI_T)-\sigma(z^*+\chi^*\chi^{*\top}-q_0I_T)||_F^2-\sum_a^T\frac{1}{2v}(\tilde\chi_a-\chi_a)^2 \nonumber
\end{align}
We define the proximals to be
\begin{align}
\mathrm{Prox}(z^*,\tilde z) = \argmax_{z\in\mathscr S(T)}\Psi(z; z^*,\tilde z)\ ,\qquad \mathrm{prox}(\chi^*,\tilde\chi,z^*,z) = \argmax_{\chi\in\mathbb R^T}\psi(\chi; \chi^*,\tilde\chi,z^*,z)\ .
\end{align}

We can now state our result on the distribution of the pre-activations on the train samples.
\begin{result}[Train samples]
\label{res:trainPreAct}
Consider $X,\xi$ being uniformly sampled from the pre-training set $\mathcal S$. Then $(z^*,z)$ follow the distribution $\mathcal R$ defined by
\begin{align}
\mathcal R : (z^*,\tilde z)\sim\mathcal Q \quad\mathrm{and}\quad z=\mathrm{Prox}(z^*,\tilde z)\ .
\end{align}
Consider $X$ being uniformly sampled from the fine-tuning set $\mathcal S'$. Then $(\chi^*,\chi, z^*, z)$ follow the distribution $\mathcal R'$ defined by
\begin{align}
\mathcal R' : (\chi^*,\tilde\chi)\sim\mathcal Q'\ ,\quad (z^*,z)\sim\left\{\begin{matrix}
\mathcal Q\quad\mathrm{if\ e}=0 \\
\mathcal R\quad\mathrm{if\ e}=1
\end{matrix}\right.
\quad\mathrm{and}\quad \chi=\mathrm{prox}(\chi^*,\tilde\chi,z^*,z)\ .
\end{align}
\end{result}

Result \ref{res:trainPreAct} shows that the pre-activation on the train samples are given by the above-introduced proximals. To give some intuition on what the proximals correspond to, we consider two limiting cases, for the pre-training. If $V\to 0$, $z=\mathrm{Prox}(z^*,\tilde z)=\tilde z$ and $(z^*,z)\sim\mathcal Q$ i.e. the pre-activations are distributed like test quantities. If $V\to\infty$, $z=\mathrm{Prox}(z^*,\tilde z)\in\sigma^{-1}(z^*)$ and the attention perfectly fits the train data. The same holds for the fine-tuning part. The proximals allow thus to interpolate between these two regimes.

A consequence of the characterizations given in results \ref{res:testPreAct} and \ref{res:trainPreAct} is that the performances of the trained attention can be entirely expressed in terms of the order parameters in Eq.~\ref{eq:order_parameters}.
\begin{result}[Performances in terms of the order parameters]
\label{res:perf}
Take $(z^*,z)\sim\mathcal Q$ and $(\chi^*,\chi)\sim\mathcal Q'$. In the high-dimensional limit the test losses of the trained attention can be expressed as
{\small
\begin{align}
\mathcal{E}(\hat W) = \mathbb E\,||\sigma(z)-\sigma(z^*)||_F^2\ , \quad \mathcal{E}'(\hat W,\hat w) = \mathbb E\,||\sigma(z+\chi\chi^\top-qI_T)-\sigma(z^*+\chi^*\chi^{*\top}-q_0I_T)||_F^2
\end{align}
}
Take $(z^*,z)\sim\mathcal R$ or $(\chi^*,\chi,z^*,z)\sim\mathcal R'$. In the high-dimensional limit the train losses of the trained attention can be expressed as
{\small
\begin{align}
\mathcal{L}(\hat W) = \mathbb E_{\mathcal R}\,||\sigma(z)-\sigma(z^*)||_F^2\ , \quad \mathcal{L}'(\hat W,\hat w) = \mathbb E_{\mathcal R'}\,||\sigma(z+\chi\chi^\top-qI_T)-\sigma(z^*+\chi^*\chi^{*\top}-q_0I_T)||_F^2
\end{align}
}
Moreover the cosine similarity is $o_w=m/\sqrt{qq_0}$.
\end{result}

\subsection{Determination of the order parameters}

The various laws $\mathcal Q, \mathcal Q', \mathcal R, \mathcal R'$ depend on the order parameters $Q, M, V, q, m$ and $v$. To fully characterize the trained attention it remains to determine them.

We introduce the spectral distribution of the pre-trained weights. Let $\mu_0$ be the limiting Marchenko-Pastur spectral density of the target $S^*$, $\mu_{\mathrm{sc},\eta}(x)=\sqrt{4\eta^2-x^2}/(2\pi\eta^2)$ the semi-circle distribution of radius $2\eta>0$, $\boxplus$ the free convolution and $\mu_\eta=\mu_0\boxplus\mu_{\mathrm{sc},\eta}$.

\begin{result}[Order parameters of the pre-training, taken from \cite{boncoraglio2026singleheadattentionhighdimensions}]
\label{res:phiPretr}
Let $\Theta=(\hat M, \hat Q, \hat V)\in\mathbb R_+^3$. Take $(z^*,\tilde z)\sim\mathcal Q$. Consider the following free entropy
\begin{align}
\Phi = -\frac{1}{2\alpha}M\hat M+\frac{1}{4\alpha}(Q\hat V-V\hat Q)+\frac{\hat M^2}{4\alpha \hat V}J\left(\frac{\hat Q^{\sfrac{1}{2}}}{\hat M},\frac{2\sqrt\kappa\lambda}{\hat M}\right) + \mathbb E_{z^*,\tilde z}\max_{z\in\mathscr S(T)}\Psi(z; z^*,\tilde z)
\end{align}
where $J(\eta,\epsilon) = \int_\epsilon^{+\infty}\mu_\eta(\mathrm dx)(x-\epsilon)^2$. 
Then $Q,M$ and $V$ are determined as $Q,M,V=\argmax(\max_\Theta\Phi)$.
\end{result}

\begin{result}[Order parameters of the fine-tuning]
\label{res:phiFT}
Let $\theta=(\hat m,\hat q,\hat v)\in\mathbb R_+^3$. Take $(z^*,z)\sim\mathcal Q$ if $\mathrm e=0$ else $(z^*,z)\sim\mathcal R$. Take $(\chi^*,\tilde\chi)\sim\mathcal Q'$. Consider the following free entropy
\begin{align}
\phi = -\frac{1}{\alpha'}m\hat m+\frac{1}{2\alpha'}(q\hat v-v\hat q)+\frac{\hat m^2+\hat q}{2\alpha'(2\lambda'+\hat v)} + \mathbb E_{\chi^*,\tilde\chi,z^*,z}\max_{\chi\in\mathbb R^T}\psi(\chi;\chi^*,\tilde\chi,z^*,z)\ .
\end{align}
Then $q,m$ and $v$ are determined as $q,m,v=\argmax(\max_\theta\phi)$.
\end{result}

Results \ref{res:phiPretr} and \ref{res:phiFT} determine the values of the order parameters in terms of the extremization of two low-dimensional potentials $\Phi$ and $\phi$. This extremization is equivalent to solving the systems of self-consistent equations described in Appendix~\ref{App:A}. Once the order parameters are determined, result~\ref{res:perf} gives the performances of the models.

\paragraph{Stability conditions.}
\label{ass:replicon}
The characterization in the above results \ref{res:trainPreAct}, \ref{res:perf}, \ref{res:phiPretr} and \ref{res:phiFT} relies on certain stability assumptions called replicon conditions. For the pre-training stage it is given in~\cite{boncoraglio2026singleheadattentionhighdimensions} and we recall it in Appendix~\ref{app:replicon}. The only new replicon condition concerns the rank-one fine-tuning stage. With $(z^*,z)\sim\mathcal Q$ if $\mathrm e=0$ and $(z^*,z)\sim\mathcal R$ if $\mathrm e=1$, and with $(\chi^*,\tilde\chi)\sim\mathcal Q'$, let $\nabla^2$ denote the Hessian of $\psi$ with respect to its first argument, evaluated at $\mathrm{prox}=\mathrm{prox}(\chi^*,\tilde\chi,z^*,z)$.
We assume that
\begin{equation}
\alpha'\,\mathbb E\left\|\left(v\nabla^2\psi(\mathrm{prox};\chi^*,\tilde\chi,z^*,z)\right)^{-1}+ I_T\right\|_F^2<1.
\label{eq:replicon2}
\end{equation}
To give some insights, the limit $v\to0$, that corresponds to a locally strongly curved fine-tuning objective, gives $v\nabla^2\psi=-I_T$ and the condition is automatically satisfied. These stability conditions do not require the original losses to be convex; see~\cite{vilucchio2025asymptotics,montanari26trivialization} for related discussions. We numerically checked that both the pre-training and fine-tuning replicon conditions are satisfied in all experiments reported in the paper.

In the following we will compare the performances of the trained attention to the \emph{Bayes-optimal} (BO) performances. These are obtained by the best achievable reconstruction of $W^*$ and $w^*$ conditionally on $\mathcal{S}, \mathcal{S'}$, assuming the knowledge of the probabilistic data model. It is the natural benchmark when having access to a limited number of samples. In the asymptotic limit we consider, the BO performances can be exactly characterized.

\begin{result}[Bayes-optimal performances]
\label{res:BO_lora_short}
In the high-dimensional limit, the BO pre-training and fine-tuning test errors $\mathcal E_\mathrm{BO}$ and $\mathcal E_\mathrm{BO}'$ are characterized by two scalar overlaps, $Q_{\rm BO}\in[0,Q_0]$ and $q_{\rm BO}\in[0,q_0]$ respectively. $Q_{\rm BO}$, $q_{\rm BO}$, $\mathcal E_\mathrm{BO}$ and $\mathcal E_\mathrm{BO}'$ are determined by the system of equations given in Appendix~\ref{app:BO_performances}.
\end{result}

\section{Consequences and applications}

\begin{figure}[t]
    \centering
    \hspace*{-1.2cm}
    \includegraphics[width=0.95\linewidth]{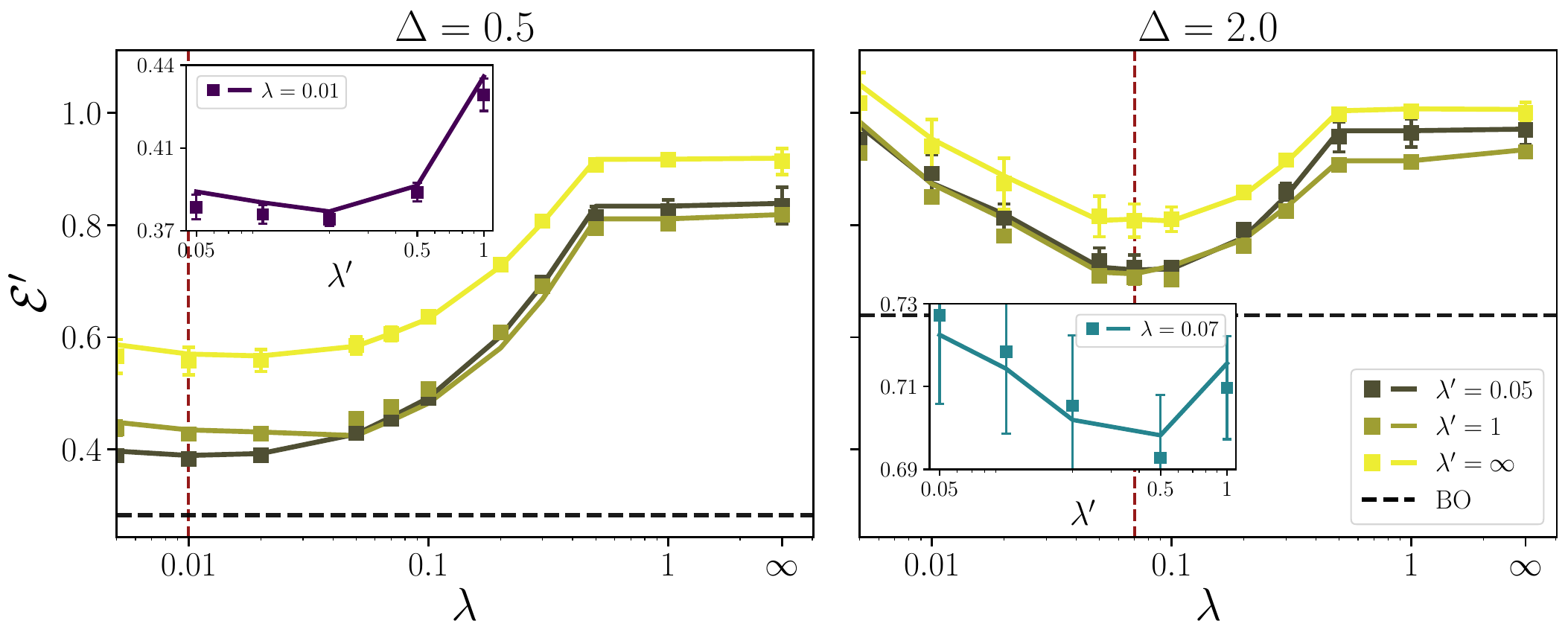}
    \caption{Fine-tuning test error $\mathcal E'$ versus pre-training regularization $\lambda$ for $\lambda'\in\{0.05,1,\infty\}$, at $\alpha=0.1$, $\alpha'=3$, $T=3$, $\kappa_0=\kappa=1$, and noise levels $\Delta=0.5$ (left) and $\Delta=2$ (right). Insets show $\mathcal E'$ versus $\lambda'$ at the optimal $\lambda$. Solid lines are state-evolution predictions; square markers are $D=150$ simulations averaged over four runs. Vertical dark-red dashed lines mark curve minima, and the black dashed line is the Bayes-optimal benchmark. The cases $\lambda'=\infty$ and $\lambda=\infty$ correspond respectively to no LoRA adaptation $(\hat w=0)$ and LoRA-only fine-tuning $(\hat W=0)$.}
    \label{fig:SE_LBFGS_check}
\end{figure}

We now discuss the main consequences of our asymptotic characterization. We start by validating the theory on the full pre-training plus LoRA pipeline with numerical simulations. We then use the theory to isolate the effects of pre-training on the fine-tuning task and we derive consequences for fine-tuning protocols.

We compare our theoretical predictions to numerical simulations of the pre-training and fine-tuning of the attention at large finite $D$ on Fig.~\ref{fig:SE_LBFGS_check} for independent fine-tuning sequences $\mathrm{e}=0$ and Fig.~\ref{fig:XXloraCorr} for shared sequences $\mathrm e=1$, showing very good agreement. Our predictions are further supported by Figs.~\ref{fig:loraTestErrVsDeltaAlpha}, and \ref{fig:activeFineTuning}, which focus on the fine-tuning given an effective pre-training. Details on the numerics are given in App.~\ref{sec:app_numerics}. 

As a baseline, we consider the two limiting cases when there is no fine-tuning or when there is no pre-training. These are obtained by respectively taking $\lambda'\to+\infty$ or $\lambda\to+\infty$. Figure~\ref{fig:SE_LBFGS_check} shows that, as expected, the LoRA consistently improves over the frozen pre-training baseline $\lambda'=\infty$, showing that the rank-one update $\hat w$ effectively exploits information not captured by $\hat W$ alone. Conversely, taking $\lambda$ too large does not produce an informative representation $\hat W$ and degrades fine-tuning test performances. Fig.~\ref{fig:SE_LBFGS_check} further informs the choice of the hyperparameters $\lambda,\lambda'$ that are optimal for the test error. 

\subsection{Independent pre-training and fine-tuning samples}

We first consider the case $\mathrm e=0$ where the pre-training and fine-tuning input samples are independent. To gain insights on the effect of the pre-training on the fine-tuning performances, we consider a linear activation function $\sigma$; we will then generalize to a softmax. The LoRA training error, expressed in terms of the pre-activations Result~\ref{res:perf}, is then
\begin{align}
\mathcal L'=T^{-2}\mathbb E||\underbrace{z-z^*}_{h}+\chi\chi^\top-\chi^*\chi^{*\top}-(q-q_0)I_T||_F^2
\end{align}
with $h_{ab}\sim\mathcal N(0,(1+\delta_{ab})\Delta_\mathrm{eff})$. The pre-training part is equivalent to an effective noise with variance
\begin{align}
\label{eq:eff_noise}
\Delta_\mathrm{eff}=\frac{\Delta}{2}+Q_0-2M+Q\ .
\end{align}
The derivation of $\Delta_\mathrm{eff}$ is straightforward and given in App.~\ref{sec:app_effectiveNoise}. The effect of pre-training is thus to reduce the noise $\Delta_\mathrm{eff}$ due to misalignment between $W^*$ and $\hat W$. By aligning the weights $\hat W$ with the ground truth, i.e. by increasing $M$, $\Delta_\mathrm{eff}$ diminishes down to the irreducible noise $\Delta/2$ in the case of a perfect learning $\hat W=W^*$.

\begin{figure}[t]
    \centering
    \includegraphics[width=0.95\linewidth]{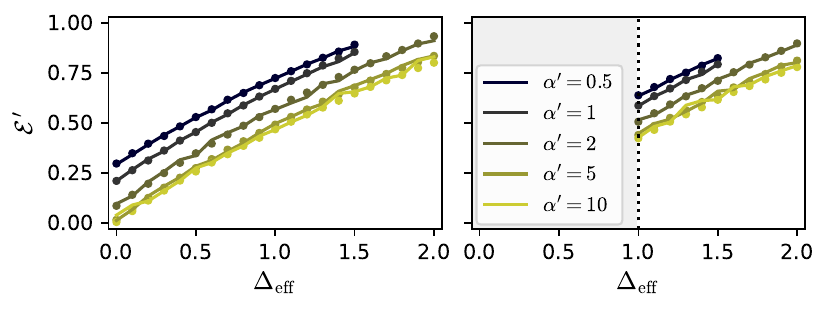}
    \caption{Effect of pre-training on the fine-tuning task. Fine-tuning test error $\mathcal E'$ vs $\Delta_\mathrm{eff}=\Delta/2+Q_0-2M+Q$, parameterized by fixing $M=1$ and varying $Q_0=Q$. Left: $\Delta=0$; right: $\Delta=2$. Lines are the theoretical predictions; dots are simulations at $D=10^4$ over five independent runs. $T=3$, $\lambda'=0.5$, $\sigma$ softmax. }
    \label{fig:loraTestErrVsDeltaAlpha}
\end{figure}

$\Delta_\mathrm{eff}$ is the value quantifying the effect of the pre-training on the fine-tuning for a linear~$\sigma$. Turning to the softmax attention, $\Delta_\mathrm{eff}$ still catches most of the pre-training effects, although we cannot reduce $z^*$ and $z$ to a unique quantity $h$. This is depicted in Fig.~\ref{fig:loraTestErrVsDeltaAlpha} where we vary $\Delta_\mathrm{eff}$ for different combinations of $\Delta, Q_0, Q$, and $M$. For the two combinations we try the curves are qualitatively the same.

Figure \ref{fig:loraTestErrVsDeltaAlpha} shows the benefits of pre-training, in terms of the number of fine-tuning samples $N'=\alpha'D$ required to achieve a given test error. In the cases we consider, diminishing $\Delta_\mathrm{eff}$ with a better pre-training is always beneficial and allows to diminish the number of required samples to reach the same performance or to decrease the fine-tuning test error at constant $\alpha'$. The same holds for a linear $\sigma$, as depicted in Fig.~\ref{fig:loraTestErrVsDeltaAlpha_lin} in Appendix~\ref{sec:supplFig}.

Finally, the same effective-noise viewpoint suggests a simple recalibration of the frozen map. Rescaling $\hat W\to\sqrt{\beta}\hat W$ by an inverse temperature $\beta$ and minimizing on $\beta$ gives, for linear $\sigma$,
\begin{equation}
\label{eq:recalibration}
\Delta_{\rm eff}(\beta)
=
\frac{\Delta}{2}+Q_0+\beta^2Q-2\beta M,
\qquad
\beta_{\rm min}=\frac{M}{Q},
\qquad
\Delta_{\rm eff}(\beta_{\rm min})
=
\frac{\Delta}{2}+Q_0-\frac{M^2}{Q}.
\end{equation}
The derivation is straightforward and is given in App.~\ref{sec:app_effectiveNoise}. Thus, when the frozen representation has the right alignment but a suboptimal scale, tuning one scalar on the fine-tuning task can improve transfer without rerunning pre-training and cross-validating on $\lambda$. This is related in spirit to the explicit scale factors used in LoRA and its variants~\cite{hu2022lora,kalajdzievski2023rslora,bini2025delora}.
In Appendix~\ref{sec:supplFig} we discuss the recalibration effect and show in Fig.~\ref{fig:recalibration_two_panel} that it is possible to further decrease the fine-tuning test error over the pre-training regularization.

\subsection{Reused sequences}

\begin{figure}[t]
    \centering
    \includegraphics[width=\linewidth]{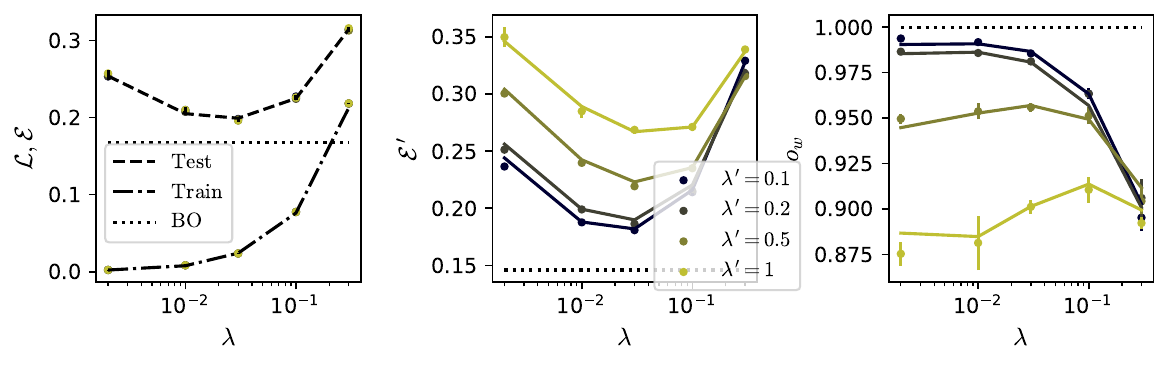}
    \caption{Fine-tuning input samples form the pre-training set $\mathrm{e}=1$. Left: pre-training train error $\mathcal L$ and test error $\mathcal E$; center: fine-tuning test error $\mathcal E'$; right: fine-tuning overlap $o_w$; vs regularizations $\lambda$ and $\lambda'$.
Lines are the theoretical predictions; dots are simulations at $D=10^3$ over five independent runs. $T=2, \Delta=0.5, \kappa_0=0.5, \kappa=1, \alpha=0.2, \alpha'=3$, $\sigma$ softmax.}
    \label{fig:XXloraCorr}
\end{figure}

We now consider the case $\mathrm e=1$ when the fine-tuning sequences are sampled from the pre-training set. In this case, the best pre-training procedure faces a trade-off between good fine-tuning test error~$\mathcal E'$ and good fine-tuning reconstruction $o_w$. This trade-off arises from the possibility to overfit and memorize the pre-train data, thus having $z_\mu\approx z^*_\mu$ for all sequences from both $\mathcal S$ and $\mathcal S'$. This reduces the effective noise during the fine-tuning to $\Delta_\mathrm{eff}(\mathrm e=1)\approx 0$ and leads to a good reconstruction of $w^*$. Because of the overfitting, the test errors $\mathcal E, \mathcal E'$ are not optimal, the gains in reconstruction being not enough to compensate. This is depicted in Fig.~\ref{fig:XXloraCorr}, which shows that, for the considered value of the model parameters, overfitting the pre-train data by taking $\lambda\to 0$ maximizes the fine-tuning overlap $o_w$; while there is a finite $\lambda\approx 0.3$ that minimizes $\mathcal E$ and $\mathcal E'$. We can provide a precise description of the reduction of the noise obtained by reusing sequences. As detailed in App.~\ref{sec:app_effectiveNoise}, for a linear $\sigma$, we have
\begin{align}
\Delta_\mathrm{eff}(\mathrm e=1)=\frac{1}{(1+4VT^{-2})^2}\Delta_\mathrm{eff}(\mathrm e=0) \label{eq:deltaEffE1}
\end{align}
where $V\geq 0$ is given by Result~\ref{res:phiPretr}. Consequently the effective noise is always smaller when reusing sequences, compared to training on new sequences at $\mathrm e=0$. When $V\to\infty$, the pre-training fits data and the effective noise goes to zero.

Our model highlights that there can be a mismatch between good test error and good reconstruction, and raises the question of what is a relevant evaluation metric. More precisely, the task defined by the fine-tuning train loss $\mathcal L'$ can be seen as a proxy for retrieving the ground truth structure $w^*$ of the data, which can be used e.g. for further downstream tasks. $w^*$ does not depend on the training procedure and it may not matter whether the model has a good test error $\mathcal E'$. A similar mismatch has been proven for unsupervised settings in~\cite{mendes2026solvable,mezard2026biased}; we extend these observations to a supervised setting.

\subsection{Active fine-tuning}

We show that our analysis provides insights on how to choose the best fine-tuning samples. In this section we assume that we have a pre-trained attention and that we have access to a larger fine-tuning dataset $\bar{\mathcal S}'$ of size $\bar N'=\bar\alpha'D$, out of which we can only label $N'=\alpha'D$ samples. We set the labeling ratio $\rho=\sfrac{N'}{\bar N'}<1$. This corresponds to the common scenario when $N'$ limited by the difficulty to label data. We seek a procedure to determine the best samples from $\bar{\mathcal S}'$ to label, compared to a uniform sampling of $\mathcal S'$ from $\bar{\mathcal S}'$. We consider the case $\mathrm e=0$ and we focus on the pre-activations of the pre-trained attention. 

We propose Algorithm \ref{alg:activeFT}, that constructs $\mathcal S'$ by choosing the sequences such that the variance over the pre-activations $z$ in Eq.~\ref{eq:preactivations} is the smallest. This procedure reduces variation in the frozen extensive-rank component, which acts as nuisance variability for the rank-one LoRA update, while leaving the fine-tuning signal unconstrained. A more detailed explanation of Alg.~\ref{alg:activeFT} is that, in the frame of our model, for linear $\sigma$, we seek to reduce the effective noise $\Delta_{\rm eff}$, namely the variance of $z-z^*$. We compute the frozen pre-activations $z_\mu$ for all $\bar N'$ candidate sequences of $\bar{\mathcal S}'$ and form $\mathcal S'$ by selecting the $N'$ sequences with the most similar values of $z_\mu$. In practice, we take the sequences for which $\sigma(z_\mu)$ is closest to $\sigma(0)$. Alg.~\ref{alg:activeFT} leads to the following effective noise. We assume that $z_\mu\approx0$ for all $X_\mu\in\mathcal S'$, which is a good approximation when the candidate pool $\bar N'$ is much larger than the labeling budget $N'$. Then for any indices $a,b$
\begin{align}
\label{eq:deltaEffActLearning}
\var_{\mathcal Q|z=0}(z-z^*)_{ab} = (1+\delta_{a,b})\Delta_{\mathrm{eff,ActFT}}\ ,\qquad \Delta_{\mathrm{eff,ActFT}}=\frac{\Delta}{2}+Q_0-\frac{M^2}{Q}
\end{align}
Details on the derivation are given in App.~\ref{sec:app_effectiveNoise}. It is immediate to see that the resulting noise $\Delta_{\mathrm{eff,ActFT}}$ is smaller than the initial noise $\Delta_{\mathrm{eff}}$ in Eq.~\ref{eq:eff_noise} for any pre-training order parameters $M,Q$.

\setlength{\columnseprule}{0.4pt}
\begin{algorithm}[h!]
\caption{Active fine-tuning.}
\label{alg:activeFT}
\begin{multicols}{2}
\textbf{Input:} $\bar N'$ sequences $X'_\mu$ from $\mathcal P'$; pre-trained attention.\\
\textbf{1.} Compute the pre-activation $z_\mu$ of the pre-trained attention for all $X'_\mu$.\\
\textbf{2.} Compute the ordering $\mu(i)$ such that $(||\sigma(z_{\mu(i)})-\sigma(0)||_F^2)_i$ is increasing.\\
\textbf{3.} Label $(X'_{\mu(i)})_{1\leq i\leq N'}$.\\
\textbf{4.} Take $\mathcal S'=(X'_{\mu(i)},y'_{\mu(i)})_{1\leq i\leq N'}$. \\
\textbf{5.} Fine-tune the attention on $\mathcal S'$.
\end{multicols}
\end{algorithm}

\begin{figure}[t!]
    \centering
    \includegraphics[width=0.95\linewidth]{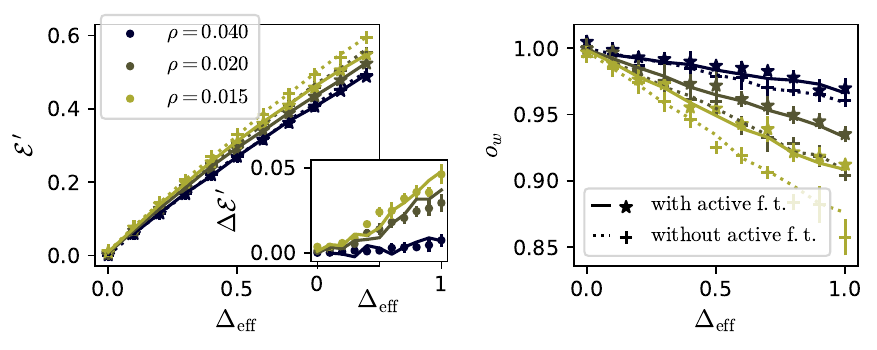}
    \caption{Active fine-tuning. Left: fine-tuning test errors $\mathcal E'(\hat W,\hat w)$ and $\mathcal E'(\hat W,\hat w_\mathrm{ActFT})$; inset: difference between the test errors $\Delta\mathcal E'$; right: overlaps $o_w$; vs $\Delta_\mathrm{eff}=\Delta/2+Q_0-2M+Q$, parameterized by fixing $M=1$ and varying $Q_0=Q$; for different labeling ratios $\rho$. Lines are the theoretical predictions; dots are simulations at $D=10^4$ over five independent runs. We minimize $\mathcal E'$ over $\lambda'\in\{0.1, 0.2, 0.5, 1\}$ for each point. $\bar\alpha'=100, T=3, \Delta=0$, $\sigma$ softmax.}
    \label{fig:activeFineTuning}
\end{figure}

We check that the benefits of the active fine-tuning procedure apply to softmax activation in Fig.~\ref{fig:activeFineTuning}. We denote as $\hat w_\mathrm{ActFT}$ the weights obtained by Alg.~\ref{alg:activeFT}, and keep the notation $\hat w$ when the $N'$ samples are uniformly chosen among the $\bar N'$ ones. We consider the difference of test error $\Delta\mathcal E'=\mathcal E'(\hat W,\hat w)-\mathcal E'(\hat W,\hat w_\mathrm{ActFT})$ as well as the overlaps $o_w$ of $\hat w_\mathrm{ActFT}$ and $\hat w$ w.r.t. the target weights $w^*$. For all noise levels we investigated, active learning improves the test error and allows for a better alignment with the ground truth. The effect is more pronounced when the effective noise $\Delta_\mathrm{eff}$ to be reduced is large, and when the labeling ratio $\rho$ is small i.e. when we have a large freedom of choice $\bar N'\gg N'$. Our theoretical results \ref{res:testPreAct} to \ref{res:phiFT} directly extend to this active fine-tuning procedure, by conditioning the distribution $\mathcal Q$ on $z,z^*$ to the desired event on $z$. The agreement with simulations depicted in Fig.~\ref{fig:activeFineTuning} is good. While being effective, Alg.~\ref{alg:activeFT} is limited to our data model; in the sense that since $z$ is almost fixed, it does not allow to explore the whole possible space of pre-activations and it needs extensions to deal with data where $z^*$ and $\chi^*$ are spread over several modes. Alg.~\ref{alg:activeFT} should be viewed as a static, analytically tractable proxy for more adaptive fine-tuning protocols, including reinforcement-learning-style settings where the examples used for adaptation are selected based on the current model behavior.

\section{Conclusions}
\label{sec:conclusions}

We introduced a solvable high-dimensional model for LoRA fine-tuning after attention pre-training, in which an extensive-rank component is learned from $N=\Theta(D^2)$ samples and a rank-one adaptation is learned from $N'=\Theta(D)$ samples. Within this setting, we obtained a sharp asymptotic characterization of the resulting predictor. Our results also exhibit regimes where test error and representation quality are misaligned, extending to supervised settings observations previously made in unsupervised learning.

The single-head, rank-one analysis presented here serves as a first step toward more general settings. A natural extension is to higher-rank LoRA updates and multi-head attention, where multiple directions or heads may compete for the information contained in the frozen representation, potentially leading to a richer structure of effective noise across tasks.

More broadly, it will be important to understand how robust this effective-noise picture is beyond the present setting. This includes extending the analysis beyond Gaussian inputs to establish universality, as well as studying learning dynamics, where the interaction between pre-training and fine-tuning unfolds over time rather than at the level of the empirical risk minimizer. The flexibility of the high-dimensional framework also makes it possible to investigate related phenomena such as catastrophic forgetting (see App.~\ref{App:catastrophic}).

\section*{Acknowledgements}

We thank E. Troiani and V. Erba for discussion on the AIM.
We acknowledge funding from the Swiss National Science Foundation grants SNSF SMArtNet (grant number 212049), and the Simons Collaboration on the Physics of Learning and Neural Computation via the Simons Foundation grant (\#1257413 (LZ)).

\bibliography{biblio}

\newpage
\appendix

\section{Derivation of the main results}
\label{App:A}
In this section we provide the derivation of the results on the asymptotic characterization of the trained attention of Section \ref{sec:characterization}. This derivation mainly relies on merging results established by previous work for the pre-training task \cite{boncoraglio2026singleheadattentionhighdimensions}, and results established for ERM of rank-one attention in \cite{cui2025high}, with an additional frozen field inherited from the pre-training. These results are based on well established techniques such as approximate message passing (AMP), Gaussian equivalence theory and replica theory from statistical physics. These techniques have been rigorously proven for similar settings in~\cite{xu2025matrixSensing,vilucchio2025asymptotics} and conjectured to extend to~\cite{boncoraglio2026singleheadattentionhighdimensions,cui2025high}.
In the present description of the results we thus focus on how these the two descriptions are merged in order to obtained the joint characterization of the pre-training and fine-tuning stages.

\subsection{Reduction to an attention-output problem}
\label{app:reduction}
We first reduce the sequence-to-sequence loss to a loss on the attention matrices. For a fixed matrix $B\in\mathbb R^{T\times T}$ and $X\sim\mathcal P_0$, one has
\begin{equation}
D^{-1}\|BX\|_F^2=\operatorname{Tr}\left(B\frac{XX^\top}{D}B^\top\right)
\xrightarrow[D\to\infty]{}\|B\|_F^2,
\end{equation}
since $XX^\top/D\to I_T$ for fixed $T$. Hence the sequence-to-sequence square loss is asymptotically equivalent to the square loss between the corresponding attention matrices. For instance,
\begin{align}
&D^{-1}\left\|\sigma(A-\mathbb E_X A)X
-\sigma(A^*-\mathbb E_X A^*+\xi)X\right\|_F^2\nonumber\\
&\hspace{2cm}=\left\|\sigma(A-\mathbb E_X A)-\sigma(A^*-\mathbb E_X A^*+\xi)\right\|_F^2+o_D(1).
\label{eq:app_seq2att}
\end{align}
We use this reduction throughout the derivation.

\subsection{Reduction to a Gaussian test channel, test samples}
\label{app:reduction_gaussian}
We show that the pre-activations of the attention are Gaussian over the test samples.
Let
\begin{equation}
S^*=P_0^{-1/2}W^*W^{*\top}, \qquad \hat S=P^{-1/2}\hat W\hat W^\top .
\end{equation}
For a fresh test sequence $X$, independent of the training sets and of $\hat W,\hat w$, define
\begin{equation}
z^*=A^*-\mathbb E_XA^*+\xi, \qquad z=A-\mathbb E_XA, \qquad
\chi^*=D^{-1/2}Xw^*, \qquad \chi=D^{-1/2}X\hat w,
\end{equation}
with
\begin{equation}
A^*=D^{-1}XS^*X^\top, \qquad A=D^{-1}X\hat S X^\top.
\end{equation}
Since $\mathbb E_X X_a^\top B X_b=\delta_{ab}\operatorname{Tr}B$ for any deterministic matrix $B$, the centering term is
\begin{equation}
(\mathbb E_X A^*)_{ab} = \delta_{ab}D^{-1}\operatorname{Tr}S^*,
\qquad (\mathbb E_X A)_{ab} = \delta_{ab}D^{-1}\operatorname{Tr}\hat S.
\end{equation}
Thus
\begin{equation}
z^*_{ab} = D^{-1}X_a^\top S^*X_b
- \delta_{ab}D^{-1}\operatorname{Tr}S^* + \xi_{ab}, \qquad z_{ab}
= D^{-1}X_a^\top \hat S X_b - \delta_{ab}D^{-1}\operatorname{Tr}\hat S .
\label{eq:app_centered_z}
\end{equation}

The vector pre-activations are Gaussian by construction:
\begin{equation}
\begin{pmatrix}
\chi^*_a\\ \chi_a
\end{pmatrix}
\sim \mathcal N\left( 0,
\begin{pmatrix}
q_0 & m\\
m & q
\end{pmatrix}
\right), \qquad
q_0=D^{-1}w^{*\top}w^*, \quad q=D^{-1}\hat w^\top\hat w, \quad
m=D^{-1}\hat w^\top w^* .
\label{eq:app_chi_test}
\end{equation}

For the quadratic pre-activations, we use the Gaussian equivalence principle for centered quadratic measurements of Gaussian data, as in~\cite{boncoraglio2026singleheadattentionhighdimensions,xu2025matrixSensing}. In the present finite-$T$ setting, this amounts to replacing the finite collection of centered matrices
\begin{equation}
X_aX_b^\top-\delta_{ab}I_D, \qquad 1\leq a\leq b\leq T,
\end{equation}
by jointly Gaussian Wigner measurements with matching covariance, up to errors vanishing as $D\to\infty$. Equivalently, for fixed $T$, the finite collection of scalar quadratic measurements is asymptotically Gaussian and its covariance is obtained by Wick contractions. In particular,
\begin{align}
D^{-2}\operatorname{Cov}_X \left(
\operatorname{Tr}S^*(X_aX_b^\top-\delta_{ab}I_D),
\operatorname{Tr}S^*(X_{a'}X_{b'}^\top-\delta_{a'b'}I_D) \right)
\nonumber\\ = \delta_{aa'}\delta_{bb'}(1+\delta_{ab}) D^{-2}\operatorname{Tr}(S^*)^2 + o_D(1),
\end{align}
and similarly for $\hat S$ and the cross-covariance between $S^*$ and $\hat S$. Including the additive noise gives, for $1\leq a\leq b\leq T$,
\begin{equation}
\label{app:eq_zstar}
\begin{pmatrix}
z^*_{ab}\\
z_{ab}
\end{pmatrix}
=
\begin{pmatrix}
z^*_{ba}\\
z_{ba}
\end{pmatrix}
\sim \mathcal Q_{ab} = \mathcal N\left( 0, (1+\delta_{ab})
\begin{pmatrix}
\Delta/2+Q_0 & M\\
M & Q
\end{pmatrix}
\right),
\end{equation}
where
\begin{equation}
Q_0=D^{-2}\operatorname{Tr}(S^*)^2, \qquad Q=D^{-2}\operatorname{Tr}\hat S^2, \qquad M=D^{-2}\operatorname{Tr}\hat S S^*.
\end{equation}
Equations~\eqref{eq:app_chi_test} and~\eqref{app:eq_zstar} give Result~\ref{res:testPreAct}. Combining them with the reduction~\eqref{eq:app_seq2att} gives the test-error part of Result~\ref{res:perf}.

\subsection{Pre-training train channel and state evolution}
\label{app:ERM_train_quantities}

The pre-training stage is unchanged with respect to~\cite{boncoraglio2026singleheadattentionhighdimensions}. We recall only the objects needed in the present paper: the train law $\mathcal R$, the free entropy $\Phi$, and the fixed-point equations used in the numerical implementation.

The train pre-activation is obtained from the proximal map
\begin{equation}
z= \operatorname{Prox}(z^*,\tilde z) = \argmax_{h\in\mathscr S(T)}
\Psi(h;z^*,\tilde z),
\label{eq:app_pretrain_prox}
\end{equation}
where $(z^*,\tilde z)\sim\mathcal Q$ and
\begin{equation}
\Psi(h;z^*,\tilde z) = -\|\sigma(h)-\sigma(z^*)\|_F^2
- \sum_{a\leq b} \frac{(\tilde z_{ab}-h_{ab})^2}{2V(1+\delta_{ab})}.
\label{eq:app_pretrain_potential}
\end{equation}
The law $\mathcal R$ is defined by
\begin{equation}
\mathcal R: \qquad (z^*,\tilde z)\sim\mathcal Q,
\qquad z=\operatorname{Prox}(z^*,\tilde z).
\label{eq:app_pretrain_R}
\end{equation}
This is the train-channel statement in Result~\ref{res:trainPreAct} for the pre-training variables.

Let $\mu_0$ be the limiting spectral law of $S^*$, let $\mu_{{\rm sc},\eta}$ be the semicircle law of radius $2\eta$, set
\begin{equation}
\mu_\eta=\mu_0\boxplus\mu_{{\rm sc},\eta}, \qquad
J(\eta,\epsilon)=\int_\epsilon^{+\infty}\mu_\eta({\rm d}x)(x-\epsilon)^2 .
\label{eq:app_J}
\end{equation}
The pre-training free entropy is
\begin{align}
\Phi(Q,M,V;\hat M,\hat Q,\hat V)
&= -\frac{1}{2\alpha}M\hat M +\frac{1}{4\alpha}(Q\hat V-V\hat Q)
+\frac{\hat M^2}{4\alpha\hat V} J\left(
\frac{\hat Q^{1/2}}{\hat M}, \frac{2\sqrt{\kappa}\lambda}{\hat M}
\right) \nonumber\\
&\hspace{1.5cm}
+
\mathbb E_{z^*,\tilde z} \max_{h\in\mathscr S(T)} \Psi(h;z^*,\tilde z).
\label{eq:app_pretraining_FE}
\end{align}
Its stable extremizer gives the order parameters $(Q,M,V)$ in Result~\ref{res:phiPretr}.

\paragraph{State-evolution equations.} We now give the equivalent fixed-point equations used to solve this extremization numerically. Set
\begin{equation}
A_0=Q_0+\frac{\Delta}{2}, \qquad B=Q-\frac{M^2}{A_0}, \qquad
s_{ab}=1+\delta_{ab}.
\end{equation}
For $1\leq a\leq b\leq T$, let $g_{ab},\zeta_{ab}$ be independent standard Gaussians and write
\begin{equation}
z^*_{ab}=\sqrt{s_{ab}A_0}\,g_{ab}, \qquad \tilde z_{ab}
= \sqrt{s_{ab}} \left( \frac{M}{\sqrt{A_0}}g_{ab} + \sqrt{B}\,\zeta_{ab}
\right).
\label{eq:app_pretrain_gaussian_param}
\end{equation}
Let $h^{\rm prox}=\operatorname{Prox}(z^*,\tilde z)$ and define
\begin{equation}
\eta=\frac{\sqrt{\hat Q}}{\hat M},\qquad \epsilon=\frac{2\sqrt{\kappa}\lambda}{\hat M}, \qquad
J_1=\partial_\eta J(\eta,\epsilon), \qquad
J_2=\partial_\epsilon J(\eta,\epsilon).
\end{equation}
The output-channel equations are
\begin{align}
\hat M
&=
\frac{2\alpha}{V} \mathbb E \sum_{a\leq b} \frac{h^{\rm prox}_{ab}-\tilde z_{ab}}{\sqrt{s_{ab}}} \left( \frac{g_{ab}}{\sqrt{A_0}} -
\frac{M}{A_0\sqrt B}\zeta_{ab} \right),
\label{eq:app_pretrain_hatM}
\\
\hat Q
&=
\frac{2\alpha}{V^2} \mathbb E \sum_{a\leq b} \frac{(h^{\rm prox}_{ab}-\tilde z_{ab})^2}{s_{ab}},
\label{eq:app_pretrain_hatQ}
\\
\hat V
&=
-\frac{2\alpha}{V\sqrt B} \mathbb E \sum_{a\leq b}
\frac{h^{\rm prox}_{ab}-\tilde z_{ab}}{\sqrt{s_{ab}}} \zeta_{ab}.
\label{eq:app_pretrain_hatV}
\end{align}
The expectations are over the Gaussians in~\eqref{eq:app_pretrain_gaussian_param}; if $B=0$, the equations are understood as the limit $B\downarrow0$. The spectral input-channel equations are
\begin{align}
Q
&=
\frac{\hat M^2}{\hat V^2} J(\eta,\epsilon),
\label{eq:app_pretrain_Q}
\\
V
&=
\frac{1}{2\hat V\eta} J_1,
\label{eq:app_pretrain_V}
\\
M
&=
\frac{\hat M}{2\hat V} \left[ 2J(\eta,\epsilon)
- \eta J_1 - \epsilon J_2 \right].
\label{eq:app_pretrain_M}
\end{align}
Equations~\eqref{eq:app_pretrain_hatM}--\eqref{eq:app_pretrain_M} are the pre-training state-evolution equations used in the simulations. At convergence, the pre-training test error is
\begin{equation}
\mathcal E(\hat W) = \mathbb E_{(z^*,z)\sim\mathcal Q}
\|\sigma(z)-\sigma(z^*)\|_F^2.
\label{eq:app_pretraining_test_error}
\end{equation}

\subsection{Decoupling of the extensive-rank and rank-one components}
\label{app:decoupling}

We now explain why the LoRA vector $w$ learned during fine-tuning is described, in the high-dimensional limit, only by its overlap with the rank-one teacher $w^*$, and not by additional overlaps with the extensive-rank matrices $S^*$ and $\hat S$.

Conditionally on the pre-training stage, $S^*$ and $\hat S$ are frozen matrices with normalized traces of order one. In the Marchenko--Pastur setting considered here, their operator norms are of order $\sqrt D$ and their squared Frobenius norms are of order $D^2$. Let $B$ denote either of the centered matrices
\begin{equation}
S^*-\frac{\operatorname{Tr}S^*}{D}I_D, \qquad
\hat S-\frac{\operatorname{Tr}\hat S}{D}I_D.
\end{equation}
Let $u,v\in\mathbb R^D$ be deterministic vectors with $\|u\|_2,\|v\|_2=O(\sqrt D)$, independent of a fresh Gaussian sequence $X$. A direct Wick contraction over $X$ gives, for any token indices $a,b,c,d$,
\begin{align}
&\operatorname{Cov}_X\left(
D^{-1}X_a^\top B X_b,\, D^{-1/2}X_c^\top u\,D^{-1/2}X_d^\top v \right)
\nonumber\\
&\hspace{2cm}
= D^{-2} \left[ \delta_{ac}\delta_{bd}\,u^\top Bv
+ \delta_{ad}\delta_{bc}\,v^\top Bu \right].
\label{eq:app_mixed_cov}
\end{align}
Since $\|B\|_{\rm op}=O(\sqrt D)$ and $\|u\|_2,\|v\|_2=O(\sqrt D)$,
\begin{equation}
D^{-2}|u^\top Bv| \leq D^{-2}\|B\|_{\rm op}\|u\|_2\|v\|_2
= O(D^{-1/2}).
\end{equation}
Therefore the mixed covariance~\eqref{eq:app_mixed_cov} vanishes as $D\to\infty$. In words, the extensive-rank quadratic fields $(z^*,z)$ and the rank-one fields $(\chi^*,\chi)$ are asymptotically independent at the level of the finite-dimensional effective channel, except for the internal covariances within each block already encoded by $(Q_0,Q,M)$ and $(q_0,q,m)$.

This also rules out additional order parameters involving both $w$ and the extensive-rank matrices. For example, for any vector $w$ with $\|w\|_2=O(\sqrt D)$,
\begin{equation}
D^{-2}|w^\top S^*w| \leq D^{-2}\|S^*\|_{\rm op}\|w\|_2^2
= O(D^{-1/2}),
\end{equation}
and similarly for $\hat S$. Hence quantities such as
\begin{equation}
D^{-2}w^\top S^*w, \qquad D^{-2}w^\top \hat S w,
\qquad D^{-2}w^{*\top}\hat S w
\end{equation}
vanish in the normalization relevant for the extensive-rank pre-activations. The rank-one LoRA component can therefore learn the direction $w^*$ at the scale $N'=\Theta(D)$, but it cannot learn the extensive-rank component, which remains a frozen Gaussian side field.

Consequently, conditionally on the pre-training order parameters $(Q,M,V)$, the fine-tuning problem reduces to a rank-one generalized linear model with teacher vector $w^*$, trainable vector $w$, and sample-wise side information $(z^*,z)$. For fresh fine-tuning samples, $(z^*,z)$ has the test law $\mathcal Q$. For fine-tuning samples reused from the pre-training set, the same decoupling holds, but $(z^*,z)$ has the train law $\mathcal R$ inherited from the pre-training stage.

\subsection{Fine-tuning free entropy and state evolution}
\label{app:finetuning_SE}

We now turn to the rank-one fine-tuning stage. Without the frozen fields $(z^*,z)$, this is a rank-one sequence multi-index ERM problem. We use the corresponding high-dimensional characterization of~\cite{cui2024highdimensionallearning,cui2025high,duranthon25slr}, with the modification that the output channel is conditioned on the pre-training fields $(z^*,z)$. By the decoupling argument of Appendix~\ref{app:decoupling}, these fields enter as side information and no additional overlaps with $S^*$ or $\hat S$ are needed.

Conditionally on the pre-training stage,
\begin{equation}
(z^*,z)\sim
\begin{cases}
\mathcal Q, & \mathrm e=0,\\
\mathcal R, & \mathrm e=1.
\end{cases}
\label{eq:app_z_law_ft}
\end{equation}
The rank-one channel is parametrized by
\begin{equation}
\chi^*\sim\mathcal N(0,I_T), \qquad \zeta\sim\mathcal N(0,I_T),
\qquad \tilde\chi=m\chi^*+\sqrt{q-m^2}\,\zeta,
\label{eq:app_chi_gaussian_param}
\end{equation}
with $\chi^*$ and $\zeta$ independent. For fixed $(\chi^*,\tilde\chi,z^*,z)$, define
\begin{align}
\psi(\chi;\chi^*,\tilde\chi,z^*,z)
&=
-
\left\| \sigma\left(z+\chi\chi^\top-qI_T\right)
- \sigma\left(z^*+\chi^*\chi^{*\top}-I_T\right)
\right\|_F^2 - \sum_{a=1}^T \frac{(\tilde\chi_a-\chi_a)^2}{2v},
\label{eq:app_ft_potential}
\end{align}
and
\begin{equation}
\chi^{\rm prox}
= \operatorname{prox}(\chi^*,\tilde\chi,z^*,z) =
\argmax_{\chi\in\mathbb R^T} \psi(\chi;\chi^*,\tilde\chi,z^*,z).
\label{eq:app_chi_prox}
\end{equation}
The fine-tuning free entropy is
\begin{align}
\phi(m,q,v;\hat m,\hat q,\hat v)
&= -\frac{1}{\alpha'}m\hat m
+\frac{1}{2\alpha'}(q\hat v-v\hat q) +\frac{\hat m^2+\hat q}{2\alpha'(2\lambda'+\hat v)} \nonumber\\
&\hspace{1.2cm}
+ \mathbb E_{\chi^*,\zeta,z^*,z}
\max_{\chi\in\mathbb R^T} \psi(\chi;\chi^*,\tilde\chi,z^*,z),
\label{eq:app_ft_free_entropy}
\end{align}
which is the formula stated in Result~\ref{res:phiFT}. The three terms in the free entropy above have the standard generalized-linear-model interpretation. The first two terms enforce the order parameters $q,m,v$, the rational term comes from integrating out the microscopic vector $w$ under the quadratic regularization $\lambda'\|w\|_2^2$, and the last term is the effective output-channel contribution. The only difference with a standard rank-one GLM is that the output channel depends on the frozen pre-training field $(z^*,z)$.

\paragraph{State-evolution equations.}
Extremizing~\eqref{eq:app_ft_free_entropy} with respect to $(\hat m,\hat q,\hat v)$ gives the following stationarity equations used in the numerical implementation. Let
\begin{equation}
C(\chi^*,\tilde\chi,z^*,z)
= -\left[ \nabla_\chi^2 \psi(\chi^{\rm prox};\chi^*,\tilde\chi,z^*,z)
\right]^{-1}.
\label{eq:app_C_def}
\end{equation}
Then the fixed point is
\begin{align}
m &= \hat m v,
\label{eq:app_se_m}
\\
q &= (\hat m^2+\hat q)v^2,
\label{eq:app_se_q}
\\
v &= \frac{1}{2\lambda'+\hat v},
\label{eq:app_se_v}
\\
\hat m
&=
\frac{\alpha'}{v} \mathbb E\left[ \chi^{*\top}\chi^{\rm prox}
- \frac{m}{v}\operatorname{Tr}C(\chi^*,\tilde\chi,z^*,z)
\right],
\label{eq:app_se_hatm}
\\
\hat q
&=
\frac{\alpha'}{v^2} \mathbb E\left[
\|\chi^{\rm prox}-\tilde\chi\|_2^2 \right],
\label{eq:app_se_hatq}
\\
\hat v
&=
\frac{\alpha'T}{v} - \frac{\alpha'}{v^2}
\mathbb E\left[ \operatorname{Tr}C(\chi^*,\tilde\chi,z^*,z)
\right] - 2\alpha' \mathbb E\left[ \partial_q \psi(\chi^{\rm prox};\chi^*,\tilde\chi,z^*,z) \right].
\label{eq:app_se_hatv}
\end{align}
All expectations are taken over~\eqref{eq:app_z_law_ft} and~\eqref{eq:app_chi_gaussian_param}. The term $\partial_q\psi$ differentiates only the explicit dependence of~\eqref{eq:app_ft_potential} on the shift $-qI_T$; the dependence of $\chi^{\rm prox}$ on $q$ is not differentiated, by the envelope theorem.

At a fixed point, the fine-tuning test error is
\begin{equation}
\mathcal E'(\hat W,\hat w) = \mathbb E_{(z^*,z)\sim\mathcal Q,\;(\chi^*,\chi)\sim\mathcal Q'} \left\| \sigma\left(z+\chi\chi^\top-qI_T\right)
- \sigma\left(z^*+\chi^*\chi^{*\top}-I_T\right) \right\|_F^2
\label{eq:app_ft_test_error_overlap}
\end{equation}
and the overlap is given by $o_w=\frac{m}{\sqrt q}$.
For softmax activation, the proximal map~\eqref{eq:app_chi_prox} is solved numerically for each Monte Carlo sample.

\subsection{Replicon stability conditions}
\label{app:replicon}

We collect here the stability conditions associated with the two state evolutions used above. They are local stability conditions for the replica-symmetric effective description around the extremizer selected by the algorithms. We verified them numerically in all parameter regimes shown in the figures.

For the extensive-rank pre-training stage, the condition is the one derived in~\cite{boncoraglio2026singleheadattentionhighdimensions,erba25nuclear}. Let $(z^*,\tilde z)\sim\mathcal Q$, let $\nabla^2$ denote the Hessian of $\Psi$ with respect to its first argument, of dimension $L\times L$ with $L=T(T+1)/2$, and write
\begin{equation}
\mathrm{Prox}=\mathrm{Prox}(z^*,\tilde z).
\end{equation}
Then the pre-training replicon condition reads
\begin{align}
2\alpha\, \mathbb E \left\|
\left(V\nabla^2\Psi(\mathrm{Prox};z^*,\tilde z)\right)^{-1}
+ I_L \right\|_F^2 \int \mu_{\hat Q^{1/2}/\hat M}({\rm d}x)
\mu_{\hat Q^{1/2}/\hat M}({\rm d}y) \frac{(\zeta(x)-\zeta(y))^2}{\hat V^2(x-y)^2} <1,
\label{eq:app_replicon_pre}
\end{align}
where
\begin{equation}
\zeta(x) = \left( x-\frac{2\sqrt{\kappa}\lambda}{\hat M} \right)_+ .
\end{equation}

The rank-one fine-tuning replicon condition is the one stated in main text. We recall it. Let $(z^*,z)\sim\mathcal Q$ if $\mathrm e=0$ and $(z^*,z)\sim\mathcal R$ if $\mathrm e=1$, let $(\chi^*,\tilde\chi)$ be distributed as in~\eqref{eq:app_chi_gaussian_param}, and write
\begin{equation}
\mathrm{prox} = \mathrm{prox}(\chi^*,\tilde\chi,z^*,z).
\end{equation}
Then the fine-tuning replicon condition is
\begin{align}
\alpha'\,\mathbb E\,\left\|(v\nabla^2\psi(\mathrm{prox};\chi^*,\tilde\chi,z^*,z))^{-1}+I_T\right\|_F^2<1\ .
\label{eq:app_replicon_finetuning}
\end{align}

The derivation is done by considering the approximate message passing (AMP) algorithm associated to the distribution on $w$ induced by the fine-tuning loss $\mathcal L'$. The AMP is stated in \cite{cui2025high}, for a generic class of sequence multi-index models. We specialize it to our case for the indices $\chi,\chi^*$ and take in account the supplementary noise $z,z^*$ from the pre-training. The replicon condition is then obtained by considering a linear perturbation of the AMP around its fixed-point and checking whether it is amplified or not. The computation follows steps similar to \cite{erba25nuclear}.

\subsection{Limiting baselines used in the figures}
\label{app:limiting_baselines}

We briefly record the two limiting baselines shown in the numerical figures.

First, when $\lambda'\to+\infty$, the LoRA update is suppressed and $\hat w=0$. Equivalently, $m,q,v\to0$ in the fine-tuning state evolution. The fine-tuning test error reduces to the frozen-pretraining baseline
\begin{equation}
\mathcal E'_{\rm frozen}(\hat W) =
\mathbb E_{(z^*,z)\sim\mathcal Q,\;\chi^*\sim\mathcal N(0,I_T)}
\left\| \sigma(z) - \sigma\left(z^*+\chi^*\chi^{*\top}-I_T\right)
\right\|_F^2 .
\label{eq:app_frozen_baseline}
\end{equation}
For the rescaled identity activation $\sigma(x)=x/T$, this becomes
\begin{equation}
\mathcal E'_{\rm frozen}(\hat W) = \frac{(T^2+T-2)\Delta_{\rm eff}+T^2+T}{T^2}, \qquad \Delta_{\rm eff} = \frac{\Delta}{2}+Q_0+Q-2M.
\label{eq:app_frozen_baseline_linear_closed}
\end{equation}

Second, when $\lambda\to+\infty$, the pre-trained extensive-rank component is suppressed and $\hat W=0$. Thus $Q=M=V=0$, $z=0$, and
\begin{equation}
z^*_{ab} \sim \mathcal N\left( 0, (1+\delta_{ab}) \left(Q_0+\frac{\Delta}{2}\right) \right).
\label{eq:app_lora_only_zstar}
\end{equation}
The fine-tuning state evolution is therefore~\eqref{eq:app_se_m}--\eqref{eq:app_se_hatv} with $z=0$ and $z^*$ distributed as in~\eqref{eq:app_lora_only_zstar}. The corresponding LoRA-only test error is
\begin{equation}
\mathcal E'_{\rm LoRA-only} = \mathbb E_{z^*,\chi^*,\chi}
\left\| \sigma\left(\chi\chi^\top-qI_T\right) -
\sigma\left(z^*+\chi^*\chi^{*\top}-I_T\right) \right\|_F^2,
\label{eq:app_lora_only_error}
\end{equation}
with $\chi^*\sim\mathcal N(0,I_T)$, $\chi\sim\mathcal N(0,qI_T)$ and $\mathbb E[\chi^*\chi^\top]=mI_T$, evaluated at the reduced fixed point.

\subsection{Catastrophic forgetting: performances of the fine-tuned model on the pre-training data}
\label{App:catastrophic}
One can evaluate the performances of the fine-tuned model on the original pre-training loss. Considering a new test pre-training sample $X,y$, it reads
\begin{equation}
\mathcal{E}''(\hat W, \hat w) = \mathbb{E}_{(X,y)\sim\mathcal P}\left[D^{-1}||\hat y(X;\hat W,\hat w)-y||_F^2 \,\big|\, \mathcal{S}, \mathcal{S}' \right]
\end{equation}
In the frame of our theory, this quantity admits an asymptotic characterization, similarly to Result~\ref{res:perf}. Taking $(z^*,z)\sim\mathcal Q$ and $(\chi^*,\chi)\sim\mathcal Q'$ we have
\begin{equation}
\mathcal{E}''(\hat W, \hat w) = \mathbb E\,||\sigma(z+\chi\chi^\top-qI_T)-\sigma(z^*)||_F^2\ .
\end{equation}
We expect this error $\mathcal{E}''$ to be higher than the pre-training test error $\mathcal E'$ because of the supplementary learned term $\chi\chi^\top-qI_T$ that acts like noise. This is reminiscent of the catastrophic forgetting phenomenon.

\section{Bayes-optimal performances}
\label{app:BO_performances}

In this appendix we give the Bayes-optimal benchmarks used in the main text. The extensive-rank pre-training BO benchmark is the single-layer AIM result of~\cite{boncoraglio2025bayes}. The fine-tuning BO benchmark follows the Bayes-optimal theory for sequence multi-index models~\cite{troiani2025fundamental}, applied to the rank-one LoRA component and conditioned on the effective pre-training channel. This combination is justified by the same decoupling mechanism as in Appendix~\ref{app:decoupling}: with only $N'=\Theta(D)$ fine-tuning samples, it is information-theoretically impossible to recover a finite fraction of the extensive-rank matrices $W^*$ or $\hat W$, whose estimation requires a quadratic number of samples. Thus the fine-tuning BO problem only estimates the rank-one direction $w^*$, while the extensive-rank part enters as a Gaussian side field determined by the BO pre-training overlap.


\subsection{Bayes-optimal pre-training for the extensive-rank AIM}
\label{app:BO_AIM_only}

We recall the single-layer BO characterization of \cite{boncoraglio2025bayes} in the notation of the present paper. Let $Q_{\rm BO}\in[0,Q_0]$ denote the BO overlap with the extensive-rank target $S^*$, and let $\hat Q_{\rm BO}\geq0$ be its conjugate. For a general output channel, the BO free entropy has the form
\begin{equation}
\Phi_{\rm BO}(Q,\hat Q) = -\frac{Q\hat Q}{4}
+ I_{\rm in}(\hat Q) + \alpha I_{\rm out}(Q),
\label{eq:BO_phi_general}
\end{equation}
and the BO overlap is obtained from the corresponding stable saddle point
\begin{equation}
(Q_{\rm BO},\hat Q_{\rm BO}) = \operatorname*{extr}_{Q\in[0,Q_0],\,\hat Q\geq0} \Phi_{\rm BO}(Q,\hat Q).
\label{eq:BO_extremization_general}
\end{equation}
For the rotationally invariant prior considered here, the input-channel equation can be written as
\begin{equation}
Q_{\rm BO} = Q_0 - \frac{1}{\hat Q_{\rm BO}} +
\frac{4\pi^2}{3\hat Q_{\rm BO}^{\,2}} \int \mu_{1/\hat Q_{\rm BO}}(t)^3\,{\rm d}t,
\label{eq:BO_pretraining_q_appendix}
\end{equation}
where $\mu_{1/\hat Q_{\rm BO}}$ is the limiting spectral density of the noisy matrix channel
\begin{equation}
Y=S^*+\hat Q_{\rm BO}^{-1/2}Z, \qquad Z\sim{\rm GOE}(D).
\label{eq:BO_noisy_matrix_channel}
\end{equation}
For the softmax output channel, the output equation is explicit:
\begin{equation}
\hat Q_{\rm BO} = \alpha\, \frac{T^2+T-2}{Q_0+\Delta/2-Q_{\rm BO}}.
\label{eq:BO_pretraining_hatq_appendix}
\end{equation}
Equations~\eqref{eq:BO_pretraining_q_appendix} and~\eqref{eq:BO_pretraining_hatq_appendix} determine the BO pre-training overlap.

The associated effective test channel is, for $a\leq b$
\begin{equation}
\left(\begin{smallmatrix}z^*_{ab}\\ z_{ab}\end{smallmatrix}\right) \sim \mathcal N\left(0,(1+\delta_{ab})\left(\begin{smallmatrix}Q_0+\Delta/2 & Q_{\rm BO}\\ Q_{\rm BO} & Q_{\rm BO}\end{smallmatrix}\right)\right)
\label{eq:BO_z_channel}
\end{equation}
where $z^*,z\in\mathscr S(T)$ are symmetrized. The BO pre-training test error is
\begin{equation}
\mathcal E_{\rm BO} = \mathbb E \left\|
\sigma(z)-\sigma(z^*) \right\|_F^2 .
\label{eq:BO_pretraining_error_appendix}
\end{equation}

\subsection{Bayes-optimal fine-tuning for the rank-one LoRA component}
\label{app:BO_lora_only}

We now state the fine-tuning BO benchmark, following~\cite{troiani2025fundamental}. Once $Q_{\rm BO}$ has been determined, the part of the extensive-rank teacher that is not known after pre-training is represented by the Gaussian residual in~\eqref{eq:BO_z_channel}. The fine-tuning data are then informative only about the rank-one vector $w^*$, while the extensive-rank component remains a side noise $(z^*,z)$ with law given by eq.~\ref{eq:BO_z_channel} for $\mathrm e=0$.

We first consider the case $\mathrm e=0$. Let $q_{\rm BO}\in[0,q_0]$ denote the overlap of the BO estimator with the fine-tuning target vector $w^*$. Its input-channel equation is
\begin{equation}
q_{\rm BO} = \frac{\hat q_{\rm BO}}{1+\hat q_{\rm BO}},
\label{eq:BO_lora_q_appendix}
\end{equation}
where $\hat q_{\rm BO}\geq0$ is the conjugate overlap.

To write the output-channel equation, let be
\begin{equation}
\left(\begin{smallmatrix}\chi^*_a\\ \chi_a\end{smallmatrix}\right) \sim \mathcal N\left(0,\left(\begin{smallmatrix}q_0 & q_{\rm BO}\\ q_{\rm BO} & q_{\rm BO}\end{smallmatrix}\right)\right)
\label{eq:BO_lora_chi_channel}
\end{equation}
for all $a\in[T]$. The output partition function is, conditionally on $\chi^*,\chi$:
\begin{equation}
Z_\mathrm{out}'=\int\prod_a\mathrm d\mathcal N(\tilde\chi_a;\omega_a,v_\mathrm{BO})\mathbb E_{z^*,z}\,\delta(\sigma(z+\tilde\chi\tilde\chi^\top-q_0I_T)-\sigma(z^*+\chi^*\chi^{*\top}-q_0I_T))
\end{equation}
where $\omega_a=\chi_a$, $v_\mathrm{BO}=q_0-q_\mathrm{BO}$ and $\delta$ an element-wise Dirac's delta. $z^*$ and $z$ are treated as noise in the output channel $\chi\mapsto\sigma(\chi\chi^\top-q_0I)$. Define $g_\mathrm{out}'=\nabla_\omega\log Z_\mathrm{out}'$ the output denoising function. The output-channel equation is then
\begin{equation}
\hat q_{{\rm BO}} = \alpha' \mathbb E_{\chi^*,\chi}\left\|g_{\rm out}' \right\|_2^2 .
\label{eq:BO_lora_hatq_appendix}
\end{equation}
One can explicit $Z_\mathrm{out}'$ for numerical computations. For $\sigma$ softmax, for $x,y\in\mathscr S(T)$, $\sigma(x)=\sigma(y)$ is equivalent to there is $s\in\mathbb R$ such that $x-y=s\mathds 1_H\mathds 1_H^\top$. One further has that $(z-z^*)_{ab}\sim\mathcal N(0,(1+\delta_{ab})V_{\rm BO})$ with $V_{\rm BO}=Q_0+\Delta/2-Q_{\rm BO}$. Integrating over $s$ gives
\begin{equation}
Z_\mathrm{out}'=\int\prod_a\mathrm d\mathcal N(\tilde\chi_a;\omega_a,v_\mathrm{BO})e^{-\frac{1}{4V_\mathrm{BO}}||\tilde\chi\tilde\chi^\top-\chi^*\chi^{*\top}||_F^2+\frac{1}{4T^2V_\mathrm{BO}}((\sum_a\tilde\chi_a)^2-(\sum_a\chi_a^*)^2)^2}
\end{equation}
Equations~\eqref{eq:BO_lora_q_appendix} and~\eqref{eq:BO_lora_hatq_appendix} determine $q_{{\rm BO}}$.

For $\mathrm e=1$, the fine-tuning inputs are among those observed at pre-training time, i.e. we are given both $\sigma(z^*)$ and $\sigma(z^*+\chi^*\chi^{*\top}-q_0I_T)$. Because of the invertibility of the function one has access to a noise-less $\chi^*$ and therefore can exactly infer $w^*$ whenever $\alpha'\geq 1$. This automatically implies $q_{\rm BO}=1$.

In both cases $\mathrm e=0$ or $\mathrm e=1$, the BO test error for fine-tuning is
\begin{equation}
\mathcal E_{\rm BO}' = \mathbb E \left\|
\sigma\left(z+\chi\chi^\top-q_0I_T\right) - \sigma\left(z^*+\chi^*\chi^{*\top}-q_0I_T\right)
\right\|_F^2,
\label{eq:BO_lora_error_appendix}
\end{equation}
where $(z,z^*,\chi,\chi^*)$ are distributed according to~\eqref{eq:BO_z_channel} and~\eqref{eq:BO_lora_chi_channel}. The BO overlap with $w^*$ is
\begin{equation}
o_{{\rm BO}} = \sqrt{q_{{\rm BO}}/q_0}.
\label{eq:BO_lora_overlap_appendix}
\end{equation}

\section{Details on the effective noises}
\label{sec:app_effectiveNoise}
We detail how the effective noises $\Delta_{\mathrm{eff}}$, $\Delta_{\rm eff}(\beta)$, $\Delta_{\mathrm{eff}}(\mathrm e=1)$ and $\Delta_{\mathrm{eff,ActFT}}$ are obtained in Eq.~\ref{eq:eff_noise}, Eq.~\ref{eq:recalibration}, Eq.~\ref{eq:deltaEffE1} and Eq.~\ref{eq:deltaEffActLearning}.

For $\mathrm e=0$ the pre-activations over the fine-tuning sets $\mathcal S'$ or $\bar{\mathcal S}'$ (active learning) follow $z^*,z\sim\mathcal Q$ (Res.~\ref{res:testPreAct}). $\mathcal Q$ factorizes over the token indices and is Gaussian. For some token indices $a\neq b$, the variance of $h_{ab}=z_{ab}-z_{ab}^*$ is
\begin{align}
\Delta_{\mathrm{eff}}&=\var(z_{ab}-z_{ab}^*) \\
&=\var z_{ab}+\var z_{ab}^*-2\mathrm{Cov}(z_{ab},z_{ab}^*) \\
&=Q+\frac{\Delta}{2}+Q_0-2M\ .
\end{align}
The case $a=b$ is similar up to a global factor 2.

Regarding the recalibration of the frozen map proposed in Eq.~\ref{eq:recalibration} it is easy to see from the order parameter definition that upon the rescaling $\hat W \to \sqrt{\beta} \hat W$, this corresponds to sending $Q\to \beta^2 Q$ and $M \to \beta M$, which leads to the recalibrated effective noise:
\begin{equation}
\Delta_{\rm eff}(\beta)
=
\frac{\Delta}{2}+Q_0+\beta^2Q-2\beta M,
\end{equation}
which is minimized for $\beta_{\rm min} = M/Q$, thus leading to the optimal effective noise upon recalibration:
\begin{equation}
\Delta_{\rm eff}(\beta_{\rm min})
=
\frac{\Delta}{2}+Q_0-\frac{M^2}{Q}. 
\end{equation}
stated in the main text.

For the active fine-tuning, we explicit the joint distribution of $z$ and $z^*$, for some token indices $a\neq b$, by taking independent $\zeta_1,\zeta_2\sim\mathcal N(0,1)$ and rewriting $\left(\begin{smallmatrix}z^*_{a,b} \\ z_{a,b}\end{smallmatrix}\right) \sim \mathcal Q_{ab}=\mathcal N\left(0,\left(\begin{smallmatrix}\Delta/2+Q_0 & M \\ M & Q\end{smallmatrix}\right)\right)$ as
\begin{align}
z_{ab}=\sqrt Q\zeta_1\ ,\qquad z_{ab}^*=\frac{M}{\sqrt Q}\zeta_1+\sqrt{\sfrac{\Delta}{2}+Q_0-\sfrac{M^2}{Q}}\zeta_2\ .
\end{align}
We assume that we can choose the fine-tuning set $\mathcal S'$ such that $\var z\approx 0$ on it. This means that $\var\zeta_1\approx 0$ and only remains the variance from $\zeta_2$, i.e.
\begin{align}
\var(z_{ab}-z^*_{ab}|z_{ab}\approx 0) = \frac{\Delta}{2}+Q_0-\frac{M^2}{Q}
\end{align}
which is $\Delta_{\mathrm{eff,ActFT}}$. The case $a=b$ is similar up to a global factor 2. We have $\Delta_{\mathrm{eff}}-\Delta_{\mathrm{eff,ActFT}}=(\sfrac{M}{\sqrt Q}-\sqrt Q)^2\geq 0$ and the active fine-tuning reduces the effective noise.

For the case $\mathrm e=1$, the pre-activations over the fine-tuning set $\mathcal S'$ follow $z^*,z\sim\mathcal R$ (Res.~\ref{res:trainPreAct}), i.e. $z=\mathrm{Prox}(z^*,\tilde z)$ with $z^*,\tilde z\sim\mathcal Q$. One has to compute the proximal $\mathrm{Prox}(z^*,\tilde z)$ defined as the optimizer of $\Psi$ over $\mathscr S(T)$. For linear $\sigma$ one has
\begin{align}
\Psi(z; z^*,\tilde z) &= -\frac{1}{T^2}||z-z^*||_F^2-\sum_{a\leq b}^T\frac{1}{2V(1+\delta_{a,b})}(\tilde z_{a,b}-z_{a,b})^2\ .
\end{align}
$\Psi$ is quadratic in $z$ and we obtain $\mathrm{Prox}(z^*,\tilde z)=(T^{-2}+(4V)^{-1})^{-1}(T^{-2}z^*+(4V)^{-1}\tilde z)$. Consequently the noise term is $z-z^*=(1+4VT^{-2})^{-2}(\tilde z-z^*)$ and
\begin{align}
\var_{z^*,z\sim\mathcal R}(z_{ab}-z^*_{ab}) &= \frac{1}{(1+4VT^{-2})^2}\var_{z^*,\tilde z\sim\mathcal Q}(\tilde z_{ab}-z^*_{ab}) \\
&= \frac{1}{(1+4VT^{-2})^2}(1+\delta_{a,b})\Delta_\mathrm{eff}\ .
\end{align}

\section{Details on the numerics}
\label{sec:app_numerics}
We describe the numerical procedures used to produce the theoretical predictions and the finite-$D$ simulations reported in the figures. The code used to generate the results is provided in the supplementary material.

The theoretical predictions are obtained by iterating until convergence the self-consistent equations in App.~\ref{app:ERM_train_quantities} and \ref{app:finetuning_SE}. The expectations are computed by Monte-Carlo sampling. We use $n_\mathrm{MC}\geq 3\times 10^4$ samples so the remaining sampling noise is comparable to the finite-$D$ fluctuations of the simulations. The proximals are numerically extremized for each sample. The iterations are stable and we use small damping.

For the theoretical predictions for active learning, we sample the pre-activations $z^*,z$ needed for the integration according to Algorithm~\ref{alg:activeFT}. We first draw $\sfrac{\bar\alpha'}{\alpha'}n_\mathrm{MC}$ Monte-Carlo samples of $z^*,z$ according to $\mathcal Q$, compute $\sigma(z)$ for each of them and select the $n_\mathrm{MC}$ samples with lowest $||\sigma(z)-\sigma(0)||_F^2$.

Regarding the simulations of the attention at given $D$, the losses of both stages are optimized using PyTorch L-BFGS with standard hyperparameters. The optimization from random initialization converge to the global minimum predicted by the theory in almost all the cases. In a few limited cases the optimization does not converge and we re-ran it with a different randomness. For Fig.~\ref{fig:activeFineTuning}, for the “without active fine-tuning” curves at large $\Delta_\mathrm{eff}$ and small $\lambda'$, the optimization does not converge in most of the cases and we initialize it at $w=w^*$. The test set size is proportional to $D$ and is taken as $10D$ samples. Error bars show the empirical standard deviation across random seeds. 

Running one theoretical prediction takes a few hours and a few GBs on a local CPU. Running one simulation at $D=10^3$ takes a few minutes and a few GBs on a local GPU. Overall we ran a hundred of predictions and a thousand of simulations.

\section{Supplementary figures}
\label{sec:supplFig}

\begin{figure}[h!]
    \centering
    \includegraphics[width=0.77\linewidth]{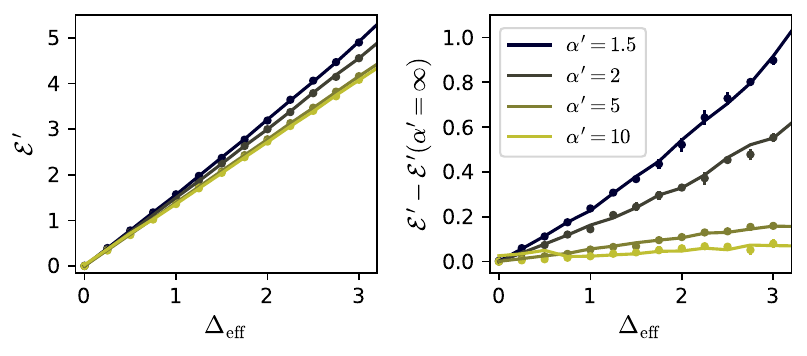}
    \caption{Effect of pre-training on the fine-tuning task for the rescaled identity activation $\sigma:x\mapsto x/T$. Fine-tuning test error $\mathcal E'$ versus $\Delta_\mathrm{eff}=\Delta/2+Q_0-2M+Q$. Left: test error; right: excess test error with respect to the value at $\Delta_\mathrm{eff}=0$. Lines are theoretical predictions; dots are simulations at $D=5\times 10^3$ over five independent runs. $T=3$.}
    \label{fig:loraTestErrVsDeltaAlpha_lin}
\end{figure}

Figure~\ref{fig:loraTestErrVsDeltaAlpha_lin} complements Fig.~\ref{fig:loraTestErrVsDeltaAlpha} by considering the rescaled identity activation $\sigma:x\mapsto x/T$. In this case the frozen pre-trained component enters the fine-tuning problem only through the scalar effective noise $\Delta_{\rm eff}=\Delta/2+Q_0-2M+Q$, as derived in App.~\ref{sec:app_effectiveNoise}. The right panel subtracts the value at $\alpha'\to\infty$ and shows the excess error due to the limited number of fine-tuning samples.

\begin{figure}[t]
\centering
\includegraphics[width=\linewidth]{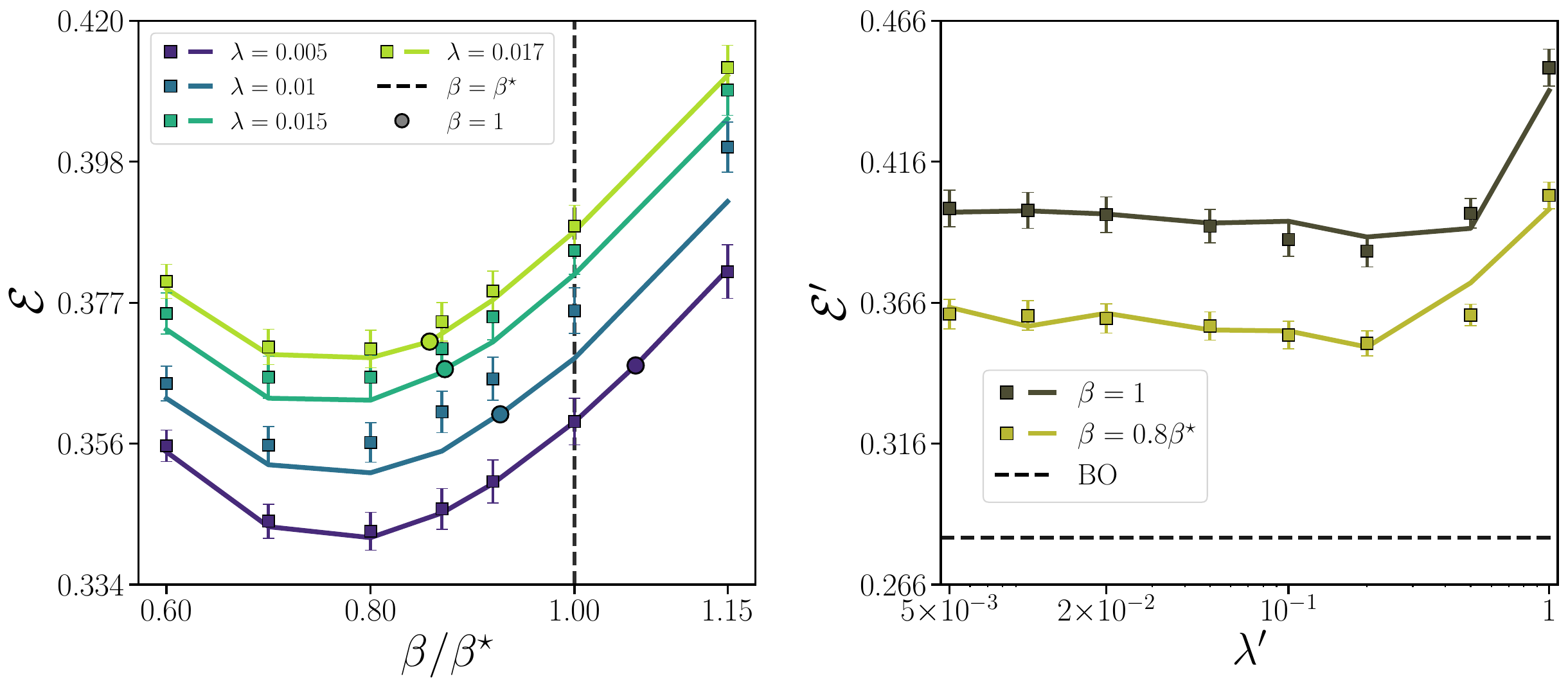}

\caption{
Recalibration of the frozen extensive-rank attention map improves downstream LoRA transfer. 
(Left:) pre-training test error $\mathcal E(\beta)$ as a function of the rescaling factor $\beta/\beta^\star$, where $\beta^\star=M/Q$. The dashed vertical line indicates $\beta=\beta^\star$, while circular markers denote the uncalibrated baseline $\beta=1$. 
(Right:) fine-tuning test error $\mathcal E'$ as a function of the LoRA regularization $\lambda'$ for the frozen representation obtained from the best pre-training configuration, i.e. $\lambda = 0.005$ and $\beta = 0.8 \beta^\star$. We compare the original frozen representation ($\beta=1$) with the recalibrated representation ($\beta=0.8\beta^\star$). Solid lines denote state-evolution predictions and square markers correspond to finite-dimensional LBFGS simulations. The horizontal dashed line in the right panel indicates the Bayes-optimal benchmark.
}
\label{fig:recalibration_two_panel}
\end{figure}

\paragraph{Finite-dimensional recalibration protocol.}
We consider the regime $\Delta=0.5,
\alpha=0.1,\alpha'=3,T=3, \kappa_0=\kappa=1$,
and study a range of pre-training regularization strengths $\lambda$. 
For each value of $\lambda$, we first train the extensive-rank pre-training estimator $\hat W$ at finite dimension using LBFGS and compute the recalibration factor $\beta^\star=\frac{M}{Q}$,
where $(Q,M)$ are the corresponding pre-training state-evolution order parameters. The frozen representation is then rescaled according to $\hat W\mapsto\sqrt{\beta}\,\hat W$,
with
\begin{equation}
\beta\in
\left\{
0.7\beta^\star,\;
0.85\beta^\star,\;
\beta^\star,\;
1.15\beta^\star,\;
1.30\beta^\star
\right\}.
\end{equation}
For each pair $(\lambda,\beta)$, we retrain the downstream LoRA vector while keeping the recalibrated matrix fixed and evaluate the corresponding generalization error. Finite-dimensional simulations are performed at $D=800$ and averaged over $10$ independent disorder realizations, with error bars indicating standard deviations. For each realization, optimization is repeated from four LBFGS initializations and the solution achieving the lowest objective value is retained. The estimation of the pre-training theoretical predictions is run for $250$ iterations with Monte Carlo output-channel sampling, and order parameters are estimated by averaging the final $30$ iterates to suppress residual sampling fluctuations. The benefit of the recalibration procedure is shown in Fig.~\ref{fig:recalibration_two_panel}.

\end{document}